\documentclass[letterpaper]{article} 
\usepackage{aaai2027}  
\usepackage[hyphens]{url}  
\usepackage{graphicx} 
\usepackage{natbib}  
\usepackage{caption} 
\usepackage{algorithm}
\usepackage{algorithmic}

\usepackage{booktabs}

\usepackage{subcaption}
\usepackage{amsmath}
\usepackage{amssymb}
\usepackage{mathtools}
\usepackage{amsthm}
\usepackage{multirow}
\usepackage{xspace}
\usepackage{pifont}

\newcommand{\cmark}{\ding{51}}%
\newcommand{\xmark}{\ding{55}}%
\newcommand{\method}{SGDS\xspace}

\begin{document}

\title{Semantic-Guided Dynamic Sparsification for Pre-Trained Model-based Class-Incremental Learning}
\author{
    Ruiqi Liu\textsuperscript{\rm 1,\rm 2},
    Boyu Diao\textsuperscript{\rm 1,\rm 2}\corresponding,
    Zijia An\textsuperscript{\rm 1,\rm 2},
    Runjie Shao\textsuperscript{\rm 1,\rm 2},
    Feixiang Ge\textsuperscript{\rm 1,\rm 2},
    Yongjun Xu\textsuperscript{\rm 1,\rm 2}
}

\affiliations{
    \textsuperscript{\rm 1}Institute of Computing Technology, Chinese Academy of Sciences, Beijing, China\\
    \textsuperscript{\rm 2}University of Chinese Academy of Sciences, Beijing, China\\
    diaoboyu2012@ict.ac.cn
}

\maketitle

\begin{abstract}
Class-Incremental Learning (CIL) requires a model to continually learn new classes without forgetting old ones. A common and efficient solution freezes a pre-trained model and employs lightweight adapters, whose parameters are often forced to be orthogonal to prevent inter-task interference. However, we argue that this parameter-constraining method is detrimental to plasticity. To this end, we propose Semantic-Guided Dynamic Sparsification (SGDS), a novel method that proactively guides the activation space by governing the orientation and rank of its subspaces through targeted sparsification. Specifically, SGDS promotes knowledge transfer by encouraging similar classes to share a compact activation subspace, while simultaneously preventing interference by assigning non-overlapping activation subspaces to dissimilar classes. By sculpting class-specific sparse subspaces in the activation space, SGDS effectively mitigates interference without imposing rigid constraints on the parameter space. Extensive experiments on various benchmark datasets demonstrate the state-of-the-art performance of SGDS.
\end{abstract}

\section{Introduction}
\label{sec:intro}

\begin{figure}[t]
    \centering
    \includegraphics[width=0.85\linewidth]{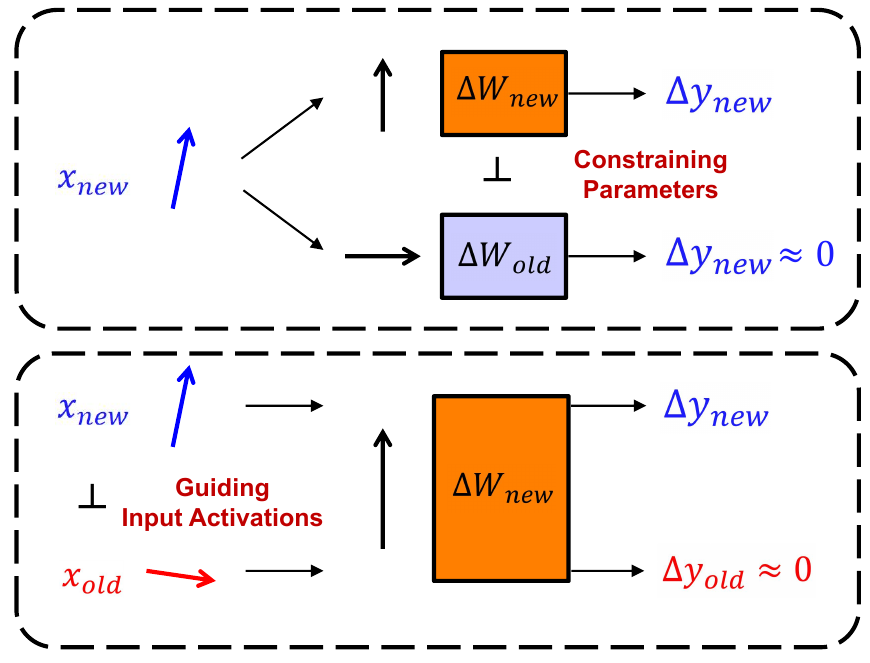} 
    \caption{
        Comparison between parameter-constraining and our activation-guiding methods.
        (a) Constraining adapter parameters (e.g., via orthogonality $\perp$) to reduce interference often harms model plasticity.
        (b) Our method guides the adapter's input activations into orthogonal subspaces. This separation mitigates inter-task interference while preserving the model's plasticity, as its parameters remain unconstrained.
    }
    \label{fig:teaser}
\end{figure}

Class-Incremental Learning (CIL) aims to enable models to continually acquire knowledge of new classes without catastrophically forgetting previously learned ones~\cite{kirkpatrick2017overcoming,mccloskey1989catastrophic}. While mitigating forgetting remains a central challenge, the increasing deployment of CIL in resource-constrained settings, such as edge devices and embodied AI, has elevated efficiency as a critical and concurrent research priority~\cite{soltoggio2024collective, liu2024resource}. 
Broadly, CIL methods can be categorized into two primary paradigms: training from scratch and leveraging Pre-trained Models (PTMs). The former paradigm, which includes regularization, architectural, or rehearsal-based methods~\cite{li2017learning, serra2018overcoming, rebuffi2017learning}, often encounters fundamental limitations concerning scalability, data privacy, or ultimate performance ceilings~\cite{wang2022dualprompt}. Consequently, the research focus has increasingly converged on CIL with PTMs~\cite{zhou2024revisiting}. This paradigm harnesses the powerful, generalizable representations learned during large-scale pre-training, providing a robust foundation that consistently leads to state-of-the-art (SOTA) performance. A prevailing method within this paradigm involves freezing the PTM backbone while employing Parameter-Efficient Fine-Tuning (PEFT) techniques, such as learning lightweight adapters~\cite{sun2025mos} or prompts~\cite{wang2022learning, smith2023coda}. This method not only excels in performance but also significantly reduces computational and storage overhead, establishing it as a dominant method for modern CIL.

A common method in Class-Incremental Learning is to constrain adapter parameters to prevent inter-task interference, often by enforcing orthogonality as shown in Figure~\ref{fig:teaser}(a). However, this parameter-constraining method harms model plasticity. Since adapters are already low-rank, further constraining their limited parameters hinders the ability to learn new tasks.
This trade-off between stability and plasticity motivates a shift from the parameter space to the activation space. Our work is grounded in the insight that inter-task interference arises not from the parameters alone, but from their interaction with neural activations. This insight suggests that instead of only constraining parameters, we can guide the activations themselves. As illustrated in Figure~\ref{fig:teaser}(b), we guide the activations for each task into orthogonal, sparse subspaces. This method mitigates interference while preserving the model's plasticity by leaving the parameters unconstrained.

To this end, we propose SGDS, a method that guides the activation space by governing the orientation and rank of its subspaces through targeted sparsification.
SGDS constructs these subspaces through a principled, two-phase process.
First, in a \textit{Semantic Exploration} phase, it controls the orientation of the activation subspaces. By assigning non-overlapping activation subspaces to dissimilar classes while encouraging shared ones for similar classes, it ensures that one subspace falls into the null space of others where appropriate.
Second, in an \textit{Activation Compaction} phase, it focuses on minimizing the rank of each subspace. By fostering sparse activations, SGDS ensures that a lower-rank activation subspace inherently creates a higher-dimensional null space~\cite{abbasi2022sparsity}. This expanded null space serves as a 'sanctuary' where activations for future tasks can be formed without interfering with the subspaces essential for prior knowledge, thereby maximizing the model's capacity to accommodate new tasks. In summary, our main contributions are as follows:

\begin{itemize}
    \item We propose SGDS, a novel method for PTM-based CIL that mitigates interference by guiding activations to form orthogonal, sparse subspaces as an alternative to constraining model parameters.
    
    \item We demonstrate that guiding the activation space provides superior performance and knowledge retention compared to directly constraining the parameter space.
    
    \item We evaluate SGDS on major CIL benchmarks. SGDS consistently outperforms SOTA methods, achieving accuracy gains of up to 1.19\%.
\end{itemize}

\section{Related Work}
\label{sec:related_work}

CIL methods can be broadly categorized into two primary paradigms: training from scratch and leveraging Pre-trained Models (PTMs).

\subsection{Training from Scratch}
The traditional paradigm of CIL trains models sequentially from scratch. These methods are typically grouped into three main families. \textbf{Regularization-based} methods~\cite{yu2020semantic,serra2018overcoming,li2017learning,yang2024clip,chaudhry2018riemannian,aljundi2018memory} introduce additional loss terms to penalize significant changes to parameters deemed critical for previous tasks. \textbf{Architecture-based} methods~\cite{ebrahimi2020adversarial,ke2020continual,loo2020generalized,wang2020learn,zhao2022deep} isolate task-specific knowledge by assigning dedicated parameters to each new task, either by expanding the network or allocating distinct sub-networks. \textbf{Rehearsal-based} methods~\cite{aljundi2019gradient,buzzega2020dark,cha2021co2l,chaudhry2021using,chaudhry2018efficient} store and replay a small subset of samples from past tasks to approximate joint training. Complementing these methods, a nascent line of work has demonstrated that enforcing sparsity on neural activations is also a potent strategy~\cite{liu2024continual,sarfraz2023sparse,abbasi2022sparsity}. By controlling which neurons are active, these methods can explicitly manage feature reuse and interference, thereby alleviating catastrophic forgetting. Despite their foundational contributions, these from-scratch methods often face fundamental limitations in scalability, data privacy, or performance ceilings, which has motivated a shift towards the PTM-based paradigm~\cite{buzzega2020dark, zhu2021class, wang2022foster, shokri2015privacy, smith2021always, wang2022dualprompt, li2024fcs}.

\subsection{Pre-Trained Model-based CIL}

To handle the high cost of training from scratch, many CIL methods now use Pre-trained Models (PTMs) with Parameter-Efficient Fine-Tuning (PEFT). This method freezes the large PTM and only fine-tunes a small set of new parameters for each task. This makes CIL practical for many resource-constrained applications. Methods in this area are diverse. Prompt-based methods learn a pool of prompts~\cite{wang2022learning,an2026parameterized}, use separate prompts for different functions~\cite{wang2022dualprompt}, or combine prompts dynamically~\cite{smith2023coda}. Other methods fine-tune parts of the backbone~\cite{zhang2023slca}, use PEFT with simple classifiers~\cite{zhou2024revisiting, mcdonnell2024ranpac}, or add new modules for each task~\cite{zhou2024expandable, sun2025mos}.

A central strategy within this paradigm, especially for adapter-based methods, is to mitigate interference at the parameter level. A representative work, TUNA~\cite{wang2025integrating}, regularizes task-specific adapters to preserve learned knowledge and combines them with a universal adapter to capture shared features. Our work SGDS uses a similar adapter framework but takes a different strategy. Instead of constraining parameters, SGDS guides the resulting activations. It guides activations for each task into sparse, separate subspaces to reduce interference. This preserves the model's parameters, better balancing stability and plasticity.

\section{Methodology}
\label{sec:methodology}

\subsection{Preliminaries}

\paragraph{Class-Incremental Learning.}
Class-Incremental Learning (CIL) sequentially learns $T$ tasks
$\{\mathcal{D}_1, \mathcal{D}_2, \ldots, \mathcal{D}_T\}$, each introducing a new set of classes. Task $t$ provides a training set
$\mathcal{D}_t = \{(x_i, y_i)\}_{i=1}^{N_t}$, where $x_i$ is an input and
$y_i \in \mathcal{C}_t$ is its label. The class sets of different tasks are disjoint, i.e.,
$\mathcal{C}_t \cap \mathcal{C}_{t'} = \emptyset$ for any $t \neq t'$.

At task $t$, only the current training set $\mathcal{D}_t$ is available for optimization. We consider exemplar-free CIL~\cite{wang2022learning,wang2022dualprompt,zhou2024revisiting}, in which no samples from previous tasks are stored. The model consists of a feature extractor $\phi$, a classifier head $W$, and trainable parameters $\theta_t$. After learning task $t$, the model is evaluated over all classes encountered so far, denoted by
$\mathcal{Y}_t = \bigcup_{i=1}^{t} \mathcal{C}_i$, without access to task identities at inference time. Therefore, CIL aims to acquire knowledge from $\mathcal{D}_t$ while preserving the ability to recognize previously learned classes in $\mathcal{Y}_{t-1}$, requiring an effective balance between plasticity and stability.

\paragraph{PTM-based CIL via Task-Specific Adapters.}
We adapt a Pre-trained Model (PTM), such as a Vision Transformer (ViT)~\cite{dosovitskiy2020image}, with lightweight modules while freezing its backbone. The trainable parameters $\theta_t$ include task-specific adapters $\mathcal{A}_t$ and a dynamic classifier head.

Following common practice~\cite{zhou2024revisiting,wang2025integrating}, a bottleneck adapter is placed in parallel with the MLP of each Transformer block. For a ViT with $L$ blocks and MLP input $\mathbf{x}_i \in \mathbb{R}^d$, it computes:
\begin{align} \label{eq:adapter_forward}
	\mathbf{x}_o = \text{MLP}(\mathbf{x}_i) + \text{ReLU}(\mathbf{x}_i W_{\text{down}})W_{\text{up}},
\end{align}
where $\mathbf{x}_o \in \mathbb{R}^d$ is the output, $W_{\text{down}} \in \mathbb{R}^{d \times r}$ is the down-projection matrix, and $W_{\text{up}} \in \mathbb{R}^{r \times d}$ is the up-projection matrix, with $r \ll d$. Task $t$ adds and trains $\mathcal{A}_t = \{W_{\text{down}}^{(l,t)}, W_{\text{up}}^{(l,t)}\}_{l=1}^L$.

Standard methods reduce inter-task interference by adding a regularization term $\mathcal{L}_{\text{reg}}$ to the classification loss. This term often imposes geometric constraints, such as orthogonality, on adapter parameters across tasks. Our method instead guides the activation space directly rather than imposing rigid parameter constraints. It therefore uses no explicit regularization term and optimizes only the standard cross-entropy loss $\mathcal{L}_{\text{cls}}$ over $\mathcal{Y}_t$:
\begin{equation}
\label{eq:cls_loss}
    \mathcal{L}_{\text{cls}} = -\frac{1}{N_t} \sum_{(\mathbf{x}, y) \in \mathcal{D}_t} \log \frac{\exp(\mathbf{w}_y^\top\phi(\mathbf{x};\mathcal{A}_t))}{\sum_{c \in \mathcal{Y}_{t}} \exp(\mathbf{w}_c^\top\phi(\mathbf{x};\mathcal{A}_t))}.
\end{equation}
Here, $\phi(\mathbf{x};\mathcal{A}_t)$ is feature extraction with adapter $\mathcal{A}_t$, and $\mathbf{w}_y$ is the classifier weight for class $y$.

At inference, the input task is unknown. A retrieval method selects $\mathcal{A}_{t^*}$ from $\{\mathcal{A}_1, \ldots, \mathcal{A}_T\}$, and the backbone $\phi$ uses it for prediction. Retrieval details are provided in Appendix.
\begin{figure*}[t]
    \centering
    \includegraphics[width=\textwidth]{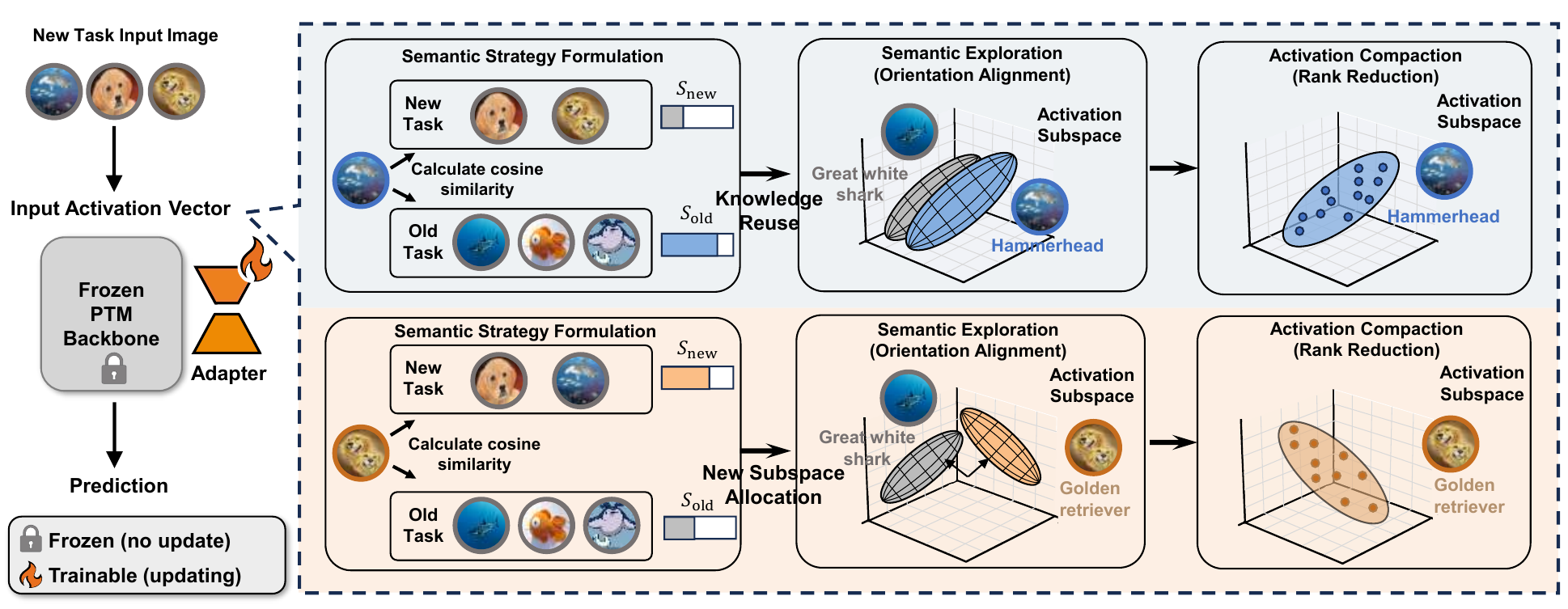} 
    \caption{\textbf{An overview of SGDS.} 
    Operating at the input of each backbone adapter, SGDS proactively guides the activation space. The process begins by selecting a learning strategy based on semantic similarity: \textbf{Knowledge Reuse} (top path) to foster knowledge transfer, or \textbf{New Subspace Allocation} (bottom path) to prevent interference. 
    This decision then guides the \textbf{Semantic Exploration} phase, which governs the subspace's orientation by either aligning it with an existing subspace or creating a new orthogonal one.
    Subsequently, the \textbf{Activation Compaction} phase reduces the subspace's rank by enforcing targeted sparsity, making it more compact. 
    This entire process delivers a strategically structured activation stream to each adapter, freeing it from inter-task interference and thereby achieving a superior stability-plasticity balance.
    }
    \label{fig:methodology}
\end{figure*}
\subsection{Our Method: Semantic-Guided Dynamic Sparsification}

SGDS constructs a well-structured null space within the activation domain by controlling two fundamental properties of activation subspaces: their orientation and rank. As illustrated in Figure~\ref{fig:methodology}, the method operates through a principled, two-phase process orchestrated by an initial semantic analysis. 
The process begins with \textit{Semantic Strategy Formulation}, which models inter-class relationships to determine whether a new class should reuse established subspaces (high similarity) or carve out new ones (low similarity). This decision then guides the two core strategies: \textbf{\textit{Semantic Exploration} and \textit{Activation Compaction}}. The \textit{Semantic Exploration} phase governs the subspace's orientation, ensuring activations for dissimilar classes are steered into non-overlapping subspaces. Subsequently, the \textit{Activation Compaction} phase controls the subspace's rank by applying targeted regularization to compact the activation representation. This induced sparsity reduces the subspace's effective rank, thereby expanding the dimension of its corresponding null space, which enhances the model's capacity to accommodate future tasks without interference.

\paragraph{Semantic Strategy Formulation.}
The first component of SGDS determines an optimal learning strategy for each new class $c \in \mathcal{C}_t$ in the incoming task $\mathcal{D}_t$. This process begins by quantifying the semantic relationship between the new classes and all known classes $\mathcal{Y}_{t-1}$. For each class $y$, we compute its prototype $\boldsymbol{\mu}_y$ by averaging the image embeddings of its training samples, extracted from the frozen PTM backbone. The relationship between a new class $c$ and any other class $y$ is then measured by the cosine similarity of their prototypes, which is converted into a probability distribution $P(y|c)$ via a softmax function, where $\text{sim}(\cdot, \cdot)$ is the cosine similarity:
\begin{equation}
\label{eq:class_relation}
P(y|c) = \frac{\exp(\text{sim}(\boldsymbol{\mu}_c, \boldsymbol{\mu}_y))}{\sum_{y' \in (\mathcal{Y}_{t-1} \cup \mathcal{C}_t)} \exp(\text{sim}(\boldsymbol{\mu}_c, \boldsymbol{\mu}_{y'}))}.
\end{equation}
This distribution informs the decision of whether class $c$ should reuse existing knowledge or be allocated new neural resources. To this end, we compute two aggregate scores: $S_{\text{old}}(c)$, which sums the probabilities connecting $c$ to all previously learned classes, and $S_{\text{new}}(c)$, which sums the probabilities connecting $c$ to the other new classes in the current task:
\begin{align}
    \label{eq:score_old}
    S_{\text{old}}(c) &= \sum_{y \in \mathcal{Y}_{t-1}} P(y|c), \\
    \label{eq:score_new}
    S_{\text{new}}(c) &= \sum_{y' \in \mathcal{C}_t} P(y'|c).
\end{align}
SGDS assigns one of two learning strategies to class $c$: if $S_{\text{old}}(c) > S_{\text{new}}(c)$, it is designated for \textit{knowledge reuse} to leverage existing features. Otherwise, it is designated for \textit{new subspace allocation} to form a distinct neural pathway.

\paragraph{Semantic Exploration.}
This phase governs the orientation of activation subspaces. 
To establish the architectural context, our method operates on a ViT with $L$ blocks. For a specific set of target blocks, indexed by a hyperparameter $\mathcal{L}_{\text{target}} \subseteq \{0, \dots, L-1\}$, SGDS sparsifies the $N$-dimensional activation vector $\mathbf{x}_l$ that serves as the input to the adapter $\mathcal{A}_l$ within each block $l \in \mathcal{L}_{\text{target}}$.
The core sparsification mechanism, applied in both the \textit{Semantic Exploration} and \textit{Activation Compaction} phases, is a two-stage process. 
First, a probabilistic sampling, modeled by a binary Bernoulli random variable $\delta_j^l$, determines the activation probability $P(\delta_j^l(c)=1)$ for the $j$-th unit in layer $l$ for a sample of class $c$.
Second, this is followed by a deterministic selection step that enforces the target sparsity: only the $\lfloor k \cdot N \rfloor$ activations with the highest magnitudes are retained, while all others are set to zero. 
This process is informed by two historical trackers, also used across both phases: a global counter $\mathbf{F} \in \mathbb{N}^{|\mathcal{L}_{\text{target}}| \times N}$ and a class-specific counter $\mathbf{F}_c \in \mathbb{N}^{|\mathcal{Y}_{t}| \times |\mathcal{L}_{\text{target}}| \times N}$.
During the \textit{Semantic Exploration} phase, the specific probability $P(\delta_j^l(c)=1)$ is determined by the learning strategy assigned to the class.
\begin{equation}
\label{eq:prob_kr}
    [p_r]_j^l(c) = 1 - \exp\left(-\sum_{y \in \mathcal{Y}_{t-1}} P(y|c) \frac{\mathbf{F}_c[y, l, j]}{\max_i(\mathbf{F}_c[y, l, i])}\right).
\end{equation}
Conversely, for classes $c$ requiring \textit{new subspace allocation}, we set $P(\delta_j^l(c)=1) = [p_a]_j^l$, where the class-agnostic probability $[p_a]_j^l$ is defined to guide activations into the null space of previously learned tasks:
\begin{equation}
\label{eq:prob_nca}
    [p_a]_j^l = \exp\left(-\beta \frac{\mathbf{F}[l, j]}{\max_i(\mathbf{F}[l, i]) }\right),
\end{equation}
where $\beta$ is a hyperparameter. The final assignment of the activation probability is formally given by:
\begin{equation}
\label{eq:final_prob}
    P(\delta_j^l(c)=1) = 
    \begin{cases} 
        [p_r]_j^l(c) & \text{if } S_{\text{old}}(c) > S_{\text{new}}(c) \\
        [p_a]_j^l & \text{otherwise}
    \end{cases}.
\end{equation}
Through this process, SGDS proactively controls the orientation of activation subspaces, either encouraging overlap for knowledge transfer or enforcing separation into null spaces to mitigate interference.

\paragraph{Activation Compaction.}
Complementing the \textit{Semantic Exploration} phase, which governs subspace orientation during the first half of the training epochs, \textit{Activation Compaction} is then applied to control the rank (i.e., size) of these subspaces. Inspired by Heterogeneous Dropout~\cite{abbasi2022sparsity}, we modulate the activation probability of each activation unit based on its historical usage for a given class. This aims to compress the activation subspace for each class $c$ into a compact and stable set of core activations, thereby preventing representational drift and enhancing stability. Specifically, for a sample of class $c$, the activation probability for the $j$-th activation in the $l$-th layer is defined as:
\begin{equation}
\label{eq:feat_consol}
    P(\delta_j^l(c)=1) =    1-\exp\left(-\gamma \frac{\mathbf{F}_c[c, l, j]}{\max_k(\mathbf{F}_c[c, l, k])}\right),
\end{equation}
where $\gamma$ is a hyperparameter modulating the compaction strength. This formulation incentivizes the model to maintain high activation magnitudes for units historically important for past classes. In summary, \textit{Semantic Exploration} first guides the subspace's orientation, while \textit{Activation Compaction} minimizes its rank. This method constructs well-structured, sparse activation subspaces tailored for each task. By proactively shaping the activation space, SGDS effectively mitigates inter-task interference, offering a more potent and flexible alternative to rigidly constraining the parameter space for achieving a superior stability-plasticity balance.
\begin{table*}[t]
\centering
\caption{
    Average ($\bar{\mathcal{A}}$) and final task ($\mathcal{A}_T$) Top-1 accuracy comparison on four challenging class-incremental learning benchmarks. 
    All methods use ViT-B/16-IN21K as the backbone and are evaluated in the exemplar-free setting.
    The specific task configurations (e.g., number of base classes and incremental steps) are detailed in the experimental setup section.
    The best performance in each column is highlighted in \textbf{bold}.
}
\label{tab:main_results}
\begin{tabular*}{\textwidth}{@{\extracolsep{\fill}}l|cc|cc|cc|cc}
    \toprule
    \multicolumn{1}{c|}{\multirow{2}{*}{\textbf{Method}}} & 
    \multicolumn{2}{c|}{\textbf{CIFAR-100}} & 
    \multicolumn{2}{c|}{\textbf{ImageNet-R}} &
    \multicolumn{2}{c|}{\textbf{ImageNet-A}} & 
    \multicolumn{2}{c}{\textbf{ObjectNet}} \\
    \cmidrule(lr){2-3} \cmidrule(lr){4-5} \cmidrule(lr){6-7} \cmidrule(lr){8-9}
    & $\bar{\mathcal{A}}$ (\%) & $\mathcal{A}_T$ (\%) & 
      $\bar{\mathcal{A}}$ (\%) & $\mathcal{A}_T$ (\%) &
      $\bar{\mathcal{A}}$ (\%) & $\mathcal{A}_T$ (\%) &
      $\bar{\mathcal{A}}$ (\%) & $\mathcal{A}_T$ (\%) \\
    \midrule
    Finetune         & 38.90 & 20.17 & 32.32 & 22.78 & 24.28 & 14.51  & 19.14 & 8.73 \\
    \midrule
    L2P~\cite{wang2022learning}            & 85.94 & 79.93 & 75.46 & 69.77 & 49.39 & 41.71 & 63.78 & 52.19 \\
    DualPrompt~\cite{wang2022dualprompt}   & 87.87 & 81.15 & 73.10 & 67.18 & 53.71 & 41.67 & 59.27 & 49.33 \\
    CODA-Prompt~\cite{smith2023coda}     & 89.11 & 81.96 & 77.97 & 72.27 & 53.54 & 42.73 & 66.07 & 53.29 \\
    SLCA~\cite{zhang2023slca}              & 92.49 & 88.55 & 81.17 & 77.00 & 68.66 & 58.74 & 72.55 & 61.30 \\ 
    SSIAT~\cite{tan2024semantically}             & 93.52 & 90.07 & 83.20 & 78.85 & 70.83 & 62.23 & 73.65 & 62.45 \\
    SimpleCIL~\cite{zhou2024revisiting}    & 87.57 & 81.26 & 61.26 & 54.55 & 59.77 & 48.91 & 65.45 & 53.59 \\
    APER + Adapter~\cite{zhou2024revisiting} & 90.65 & 85.15 & 75.82 & 67.95 & 60.47 & 49.37 & 67.18 & 55.24 \\
    RanPAC~\cite{mcdonnell2023ranpac}    & 94.00 & 90.62 & 82.98 & 77.94 & 69.32 & 61.82 & 72.76 & 62.02 \\
    EASE~\cite{zhou2024expandable}         & 91.51 & 85.80 & 81.74 & 76.17 & 65.34 & 55.04 & 70.84 & 57.86 \\
    MOS~\cite{sun2025mos}                  & 93.30 & 89.25 & 82.96 & 77.93 & 67.08 & 56.22 & 74.69 & 63.62 \\
    TUNA~\cite{wang2025integrating} & 94.44 & 90.74 & 84.22 & 79.42 & 73.78 & 64.78 & 76.46 & 66.32 \\
    \midrule
    \textbf{\method (Ours)}                   & \bf{94.98} & \bf{91.59} & \bf{85.41} & \bf{80.52} & \bf{74.39} & \bf{66.56} & \bf{76.97} & \bf{66.51} \\
    \bottomrule
\end{tabular*}
\end{table*}
\begin{figure*}[t]
    \centering
    \begin{subfigure}[b]{0.32\textwidth}
        \centering
        \includegraphics[width=\textwidth]{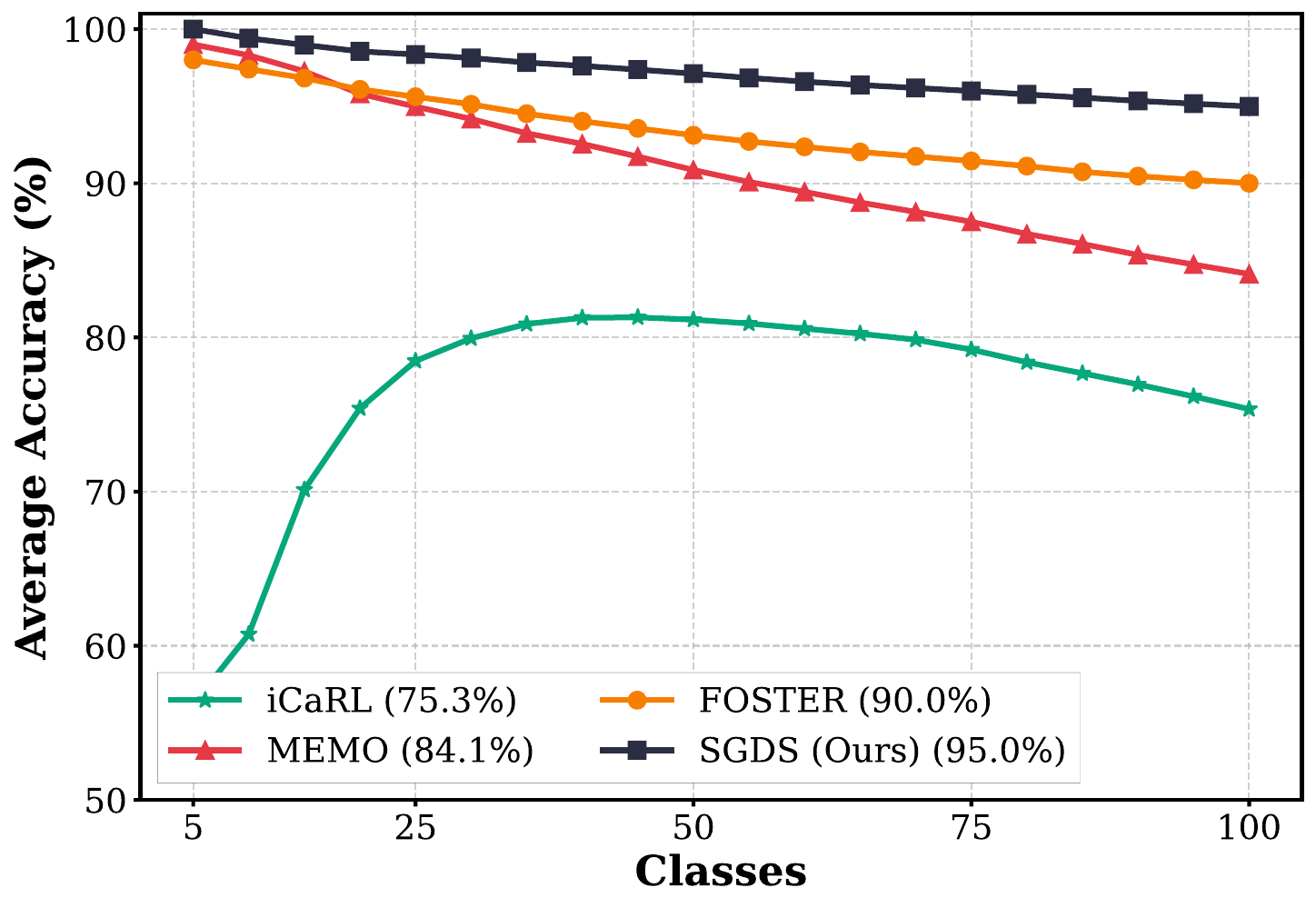}
        \caption{CIFAR-100}
        \label{fig:rehearsal_cifar}
    \end{subfigure}
    \hfill 
    \begin{subfigure}[b]{0.32\textwidth}
        \centering
        \includegraphics[width=\textwidth]{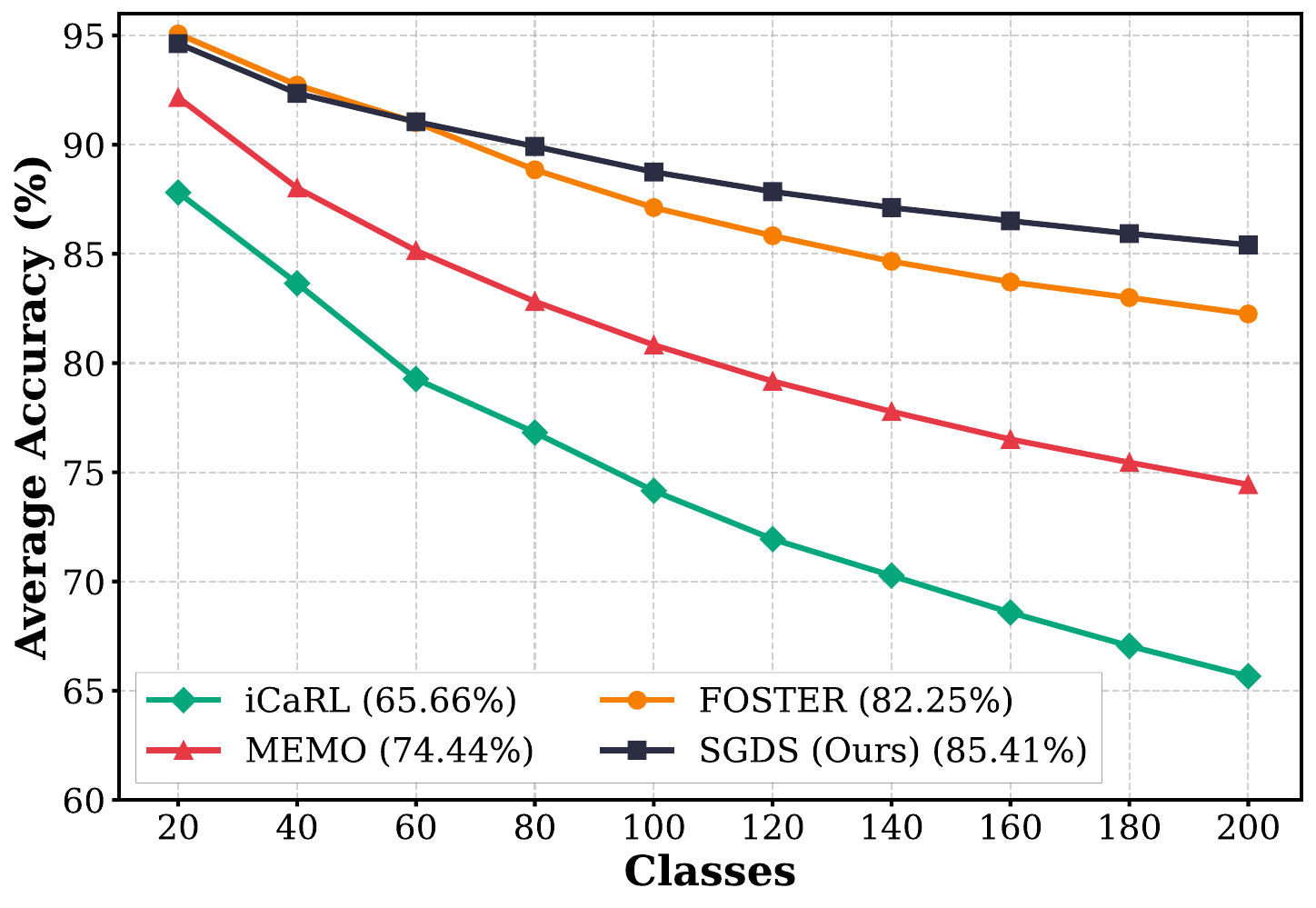}
        \caption{ImageNet-R}
        \label{fig:rehearsal_imagenetr}
    \end{subfigure}
    \hfill 
    \begin{subfigure}[b]{0.32\textwidth}
        \centering
        \includegraphics[width=\textwidth]{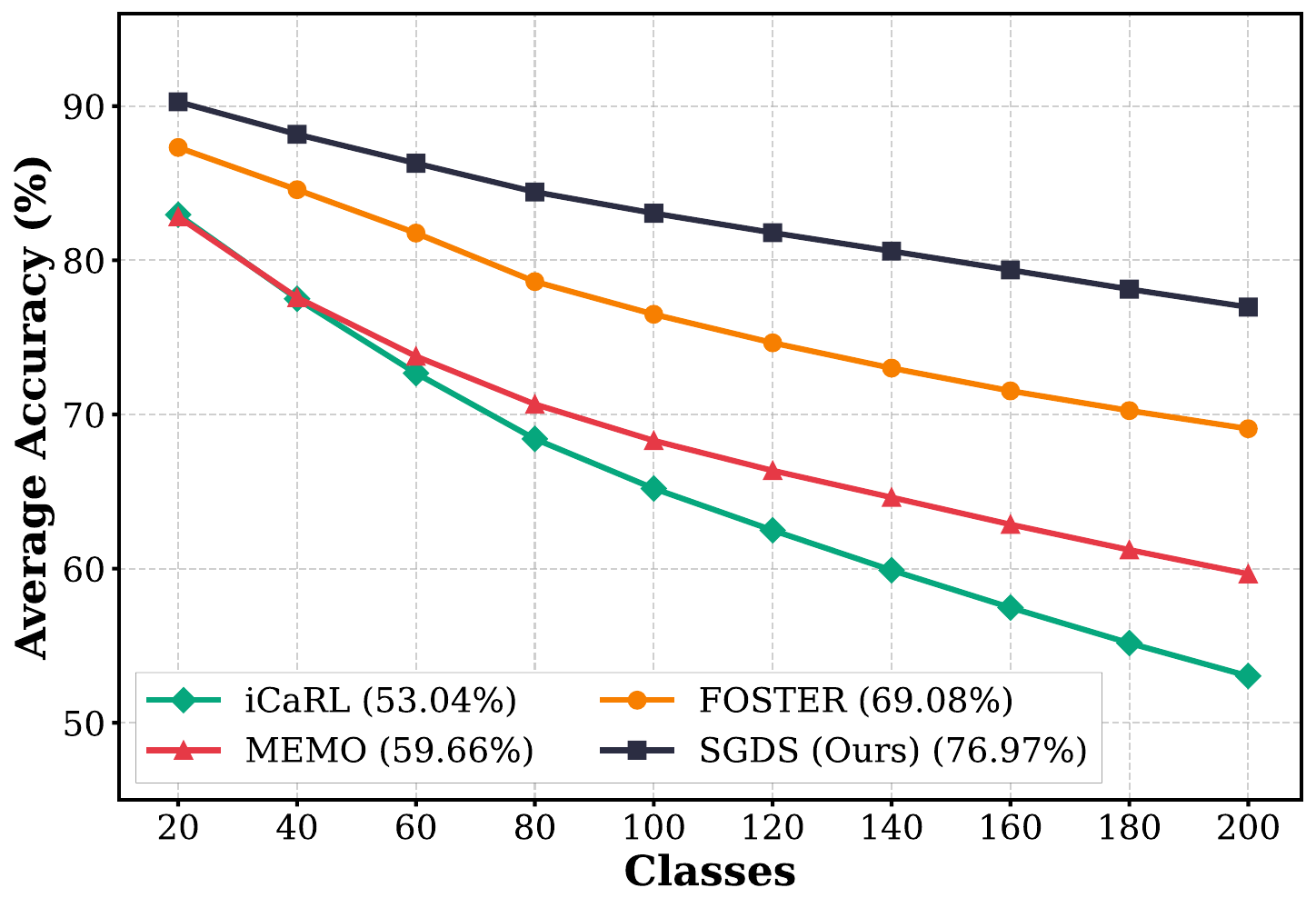} 
        \caption{ObjectNet}
        \label{fig:rehearsal_objectnet}
    \end{subfigure}
    
    \caption{
        Performance comparison with rehearsal-based methods on CIFAR-100 (left), ImageNet-R (center), and ObjectNet (right).
        The plots track the average accuracy as the model learns from a sequence of tasks. 
        SGDS consistently outperforms strong baselines, and the performance gap widens over time, indicating superior knowledge retention.
    }
    \label{fig:rehearsal_comparison}
\end{figure*}
\section{Experiments}
\label{sec:experiments}

\subsection{Implementation Details}

\noindent\textbf{Datasets and Task Protocols.}
We follow the protocol of~\cite{zhou2024revisiting} and evaluate on CIFAR-100~\cite{krizhevsky2009learning}, ImageNet-R~\cite{hendrycks2021many}, ImageNet-A~\cite{hendrycks2021natural}, and ObjectNet~\cite{barbu2019objectnet}. CIFAR-100 is split into 20 tasks with 5 classes each, while each of the other datasets is split into 10 tasks with 20 classes each. Following~\cite{rebuffi2017icarl}, we shuffle the class orders with a fixed random seed of 1993.

\noindent\textbf{Training Details.}
For a fair comparison, all methods use the same ViT-B/16 backbone pre-trained on ImageNet-21K. For each task, we train the models for 20 epochs with a batch size of 48 using SGD with momentum, an initial learning rate of 0.01, and cosine annealing. SGDS applies 60\% global activation sparsity only to the final layer of the ViT backbone. We set $\beta$ (Eq.~\ref{eq:prob_nca}) to 0.5 and $\gamma$ (Eq.~\ref{eq:feat_consol}) to 1.0. All adapter-based methods use an intermediate projection dimension of 16. Further implementation and hardware details are provided in Appendix.

\begin{figure*}[t]
    \centering
    \begin{subfigure}[b]{0.32\textwidth}
        \centering
        \includegraphics[width=\textwidth]{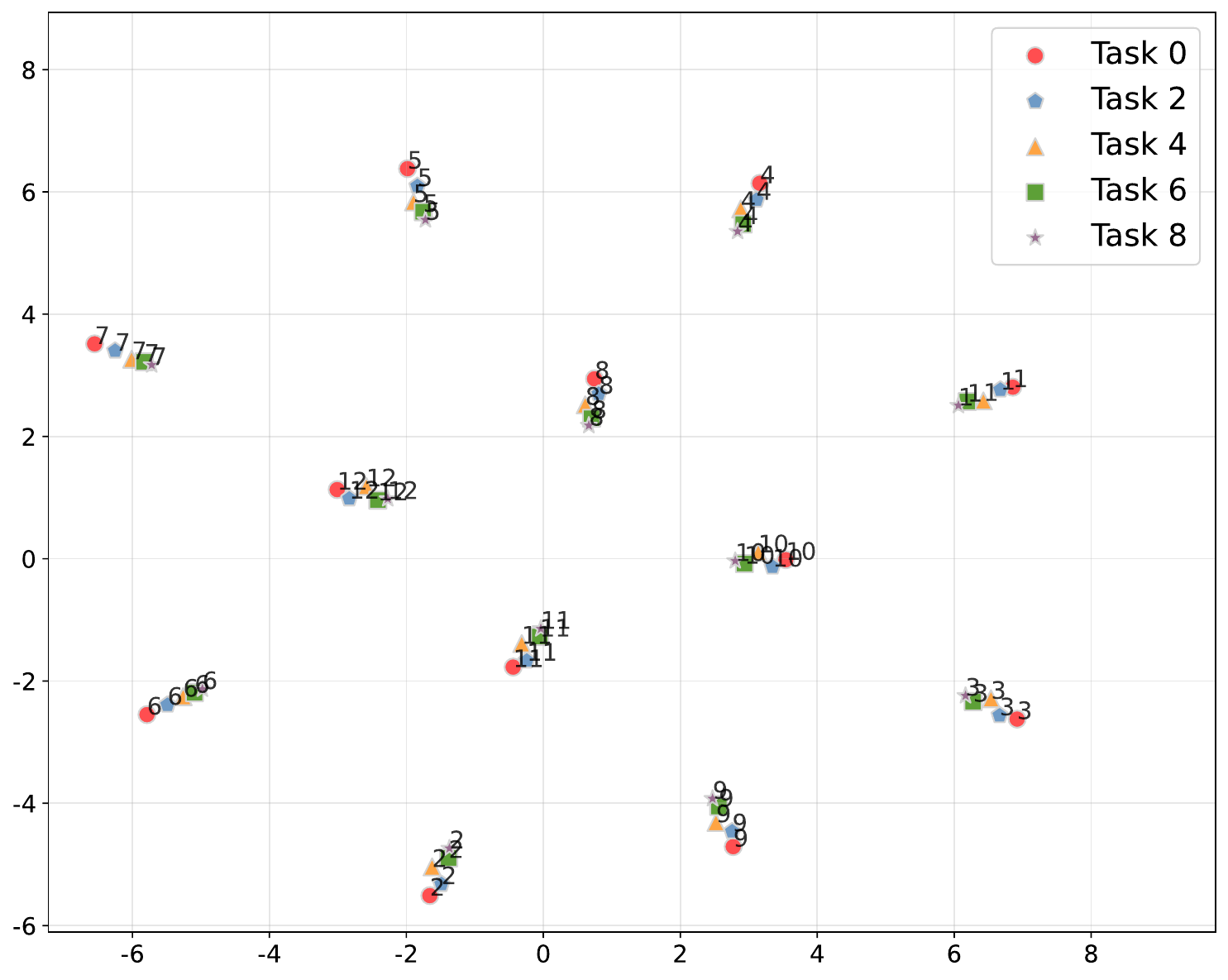} 
        \caption{Distribution of $W_{\text{down}}$ parameters.}
        \label{fig:dist_down}
    \end{subfigure}
    \hfill
    \begin{subfigure}[b]{0.32\textwidth}
        \centering
        \includegraphics[width=\textwidth]{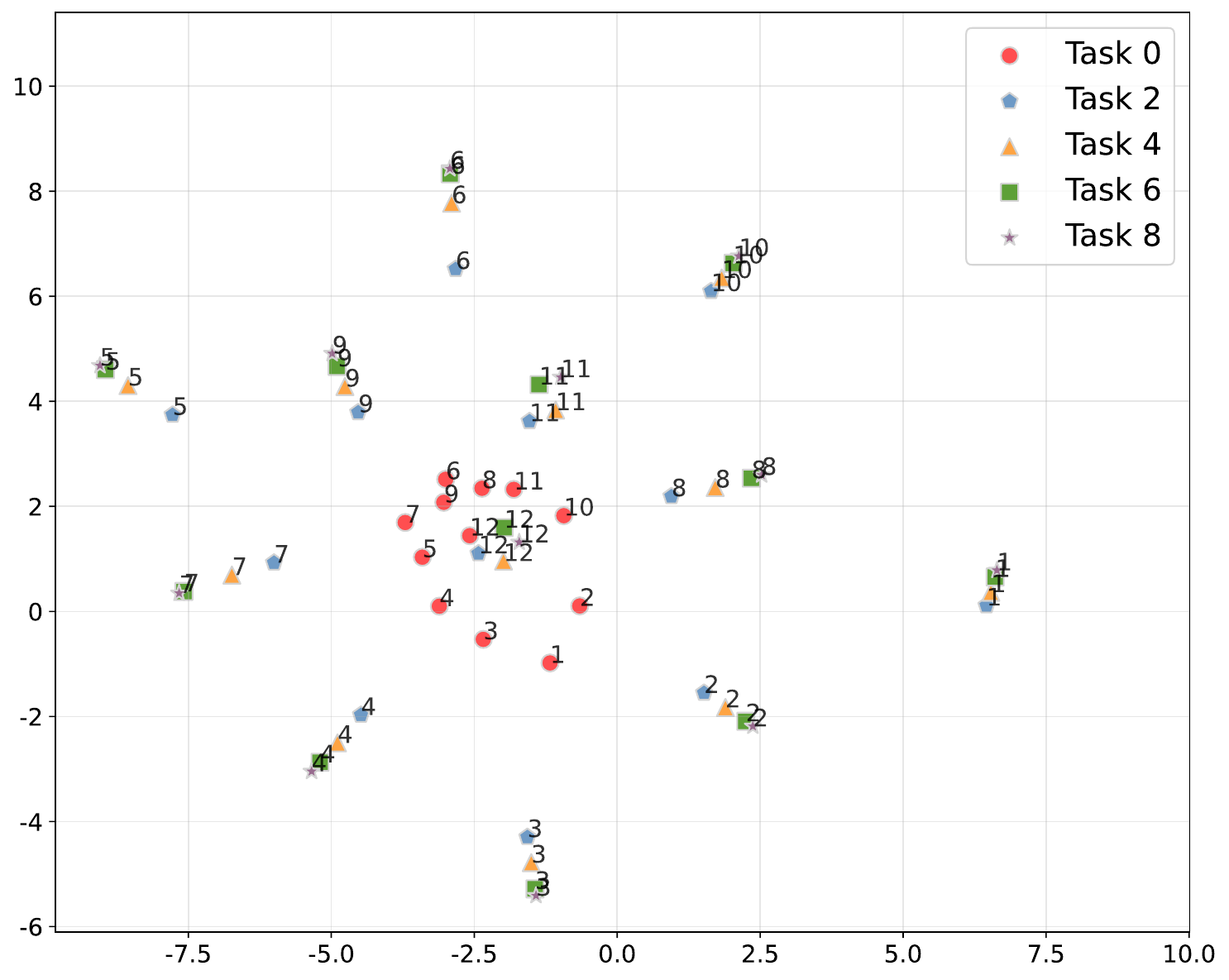} 
        \caption{Distribution of $W_{\text{up}}$ parameters.}
        \label{fig:dist_up}
    \end{subfigure}
    \hfill
    \begin{subfigure}[b]{0.32\textwidth}
        \centering
        \includegraphics[width=\textwidth]{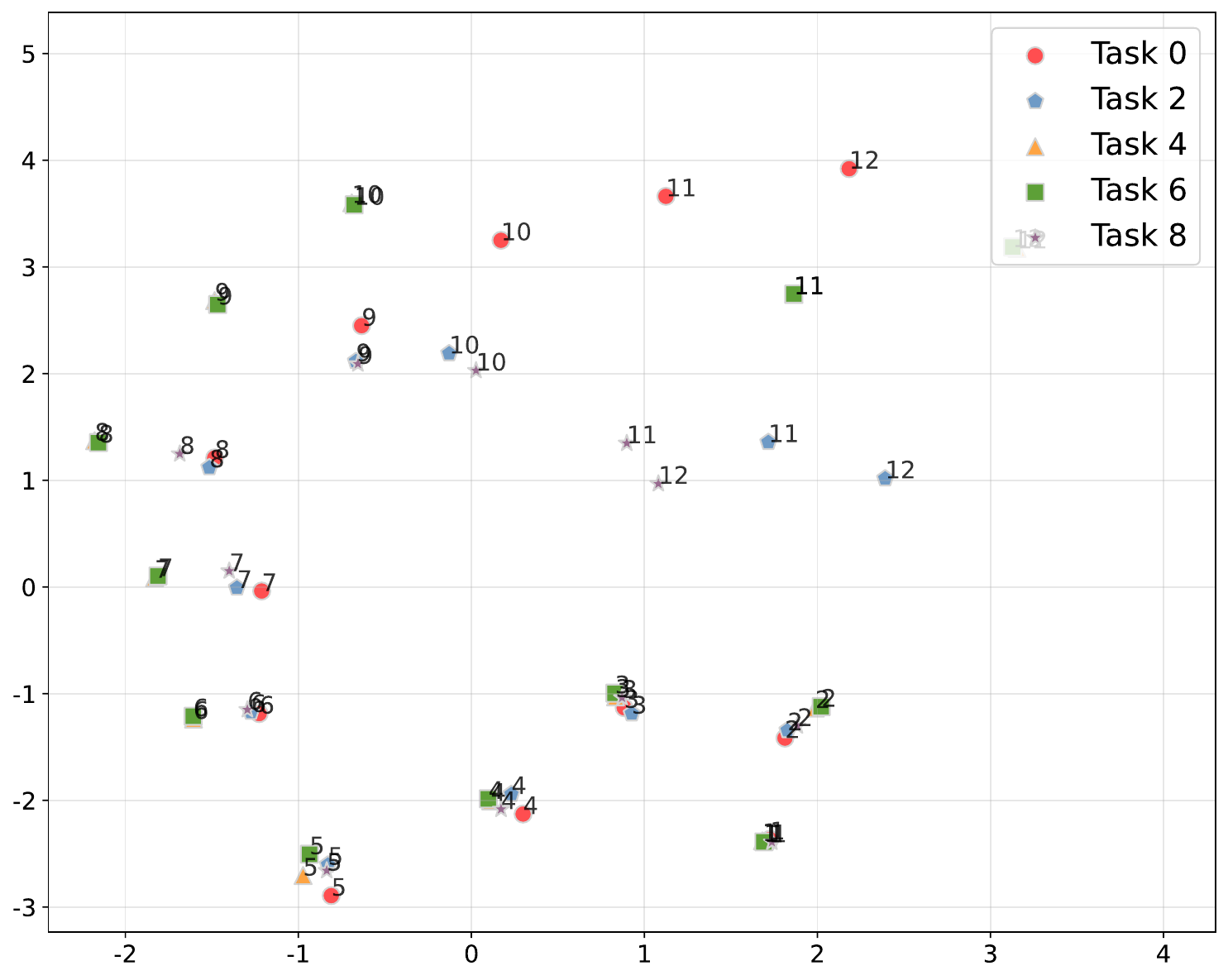} 
        \caption{Distribution of adapter input activations.}
        \label{fig:dist_features}
    \end{subfigure}
    
    \caption{
        t-SNE visualizations of parameter and activation distributions on ImageNet-R. 
        Colors represent tasks, and numbers denote layers.
        (a) The down-projection parameters ($W_{\text{down}}$) cluster tightly by layer regardless of the task, indicating they learn a shared, task-agnostic function.
        (b) In contrast, the up-projection parameters ($W_{\text{up}}$) are dispersed and form task-specific clusters, revealing their role in specializing for each task.
        (c) Similarly, the input activations are intermingled in early layers but become clearly separated by task in later layers.
    }
    \label{fig:param_vs_act}
\end{figure*}

\noindent\textbf{Comparison Methods.}
To comprehensively benchmark SGDS, we compare it against a wide array of state-of-the-art methods. The primary comparison set includes leading PTM-based CIL methods: L2P~\cite{wang2022learning}, DualPrompt~\cite{wang2022dualprompt}, CODA-Prompt~\cite{smith2023coda}, RanPAC~\cite{mcdonnell2023ranpac}, SimpleCIL~\cite{zhou2024revisiting}, APER~\cite{zhou2024revisiting}, SLCA~\cite{zhang2023slca}, SSIAT~\cite{tan2024semantically}, EASE~\cite{zhou2024expandable}, MOS~\cite{sun2025mos}, and TUNA~\cite{wang2025integrating}. We also include several prominent rehearsal-based methods adapted to the PTM setting, including FOSTER~\cite{wang2022foster}, MEMO~\cite{zhou2022model}, and iCaRL~\cite{rebuffi2017icarl}. The 'Finetune' baseline sequentially fine-tunes the entire PTM on each task. This provides a lower bound and demonstrates the severity of catastrophic forgetting.

\noindent\textbf{Evaluation Metrics.}
We follow the standard CIL evaluation protocol defined in~\cite{rebuffi2017icarl}. Performance is measured using two key metrics. The \textbf{Average Accuracy} ($\bar{\mathcal{A}}$) measures the overall performance across the entire learning trajectory: $\bar{\mathcal{A}} = \frac{1}{T} \sum_{t=1}^{T} \left( \frac{1}{t} \sum_{j=1}^{t} A_{t,j} \right)$, where $A_{t,j}$ is the accuracy on task $j$ after training on task $t$. The \textbf{Final Accuracy} ($\mathcal{A}_T$) evaluates the average performance on all tasks at the end of the sequence: $\mathcal{A}_T = \frac{1}{T} \sum_{j=1}^{T} A_{T,j}$. Higher values for both metrics signify superior CIL performance.


\subsection{Main Results}

We evaluate SGDS against SOTA exemplar-free methods, with detailed results presented in Table~\ref{tab:main_results}. 
Across all four benchmarks, SGDS consistently outperforms all competitors in both average accuracy ($\bar{\mathcal{A}}$) and final accuracy ($\mathcal{A}_T$). 
For instance, on the challenging ImageNet-R benchmark SGDS achieves an average accuracy of 85.41\%, outperforming the leading parameter-space method TUNA by a significant 1.19\%. 
This result strongly validates our core thesis that guiding the activation space is a more effective and flexible strategy for mitigating interference than constraining the parameter space. 
While parameter-space methods like TUNA are effective, their rigid constraints can limit plasticity. 
SGDS circumvents this limitation by operating dynamically in the activation domain, thus preserving parameter plasticity.

We also compare SGDS against strong rehearsal-based methods, despite our method being exemplar-free. 
As shown in Figure~\ref{fig:rehearsal_comparison}, SGDS surpasses these methods on CIFAR-100, ImageNet-R, and ObjectNet. 
On the 10-task ObjectNet sequence, SGDS achieves a final average accuracy of 76.97\%, a notable lead of 7.89\% over FOSTER. 
Notably, the widening performance gap as more tasks are learned suggests that constructing well-structured activation subspaces is a more scalable method for CIL than relying on the overhead of storing and replaying exemplars.

\begin{table}[t]
\centering
\caption{
    Ablation study on ImageNet-R. We analyze the contribution of the two core components of our proposed SGDS: \textit{Semantic Exploration} (SE) and \textit{Activation Compaction} (AC). The best performance in each column is highlighted in \textbf{bold}. The results confirm that both modules are crucial for optimal performance.
}
\label{tab:ablation_study}
\begin{tabular}{cc|cc}
\toprule
\multicolumn{1}{c}{SE} & \multicolumn{1}{c|}{AC} & \multirow{2}{*}{$\bar{\mathcal{A}}$ (\%)} & \multirow{2}{*}{$\mathcal{A}_T$ (\%)} \\ 
\multicolumn{1}{c}{\small (Exploration)} & \multicolumn{1}{c|}{\small (Compaction)} & & \\
\midrule
\xmark & \xmark & 83.42 & 79.27 \\
\xmark & \cmark & 84.53 & 79.23 \\
\cmark & \xmark & 84.66 & 79.35 \\
\cmark & \cmark & \textbf{85.41} & \textbf{80.52} \\
\bottomrule
\end{tabular}
\end{table}
\begin{table}[t]
\centering
\caption{
    Comparison of Activation Guidance and Parameter Constraints.
    We compare two distinct strategies for mitigating interference on ImageNet-R. 
    The first group applies orthogonality regularization directly to the adapter's parameter space. 
    The second group implements SGDS by guiding the activation space at different network depths.
    The results show that guiding activations, particularly in later layers, yields superior performance.
}
\label{tab:discussion_param_vs_act}
\begin{tabular}{c|cc}
\toprule
\textbf{Strategy} & $\bar{\mathcal{A}}$ (\%) & $\mathcal{A}_T$ (\%) \\
\midrule
\multicolumn{3}{c}{\textit{Parameter-Space Regularization}} \\
\midrule
\quad Regularize $W_{\text{down}}$ only & 83.33 & 79.12 \\
\quad Regularize $W_{\text{up}}$ only   & 84.37 & 79.87 \\
\quad Regularize Both ($W_{\text{down}} \& W_{\text{up}}$) & 83.22 & 78.55 \\
\midrule
\multicolumn{3}{c}{\textit{Activation-Space Guidance (Ours)}} \\
\midrule
\quad SGDS on All Layers (1-12)     & 84.59 & 80.10 \\
\quad SGDS on Last 6 Layers (7-12)  & 85.23 & 80.37 \\
\quad SGDS on Last Layer (12)       & 85.41 & 80.52 \\
\bottomrule
\end{tabular}
\end{table}
\subsection{Ablation Study}

We conduct a detailed ablation study on ImageNet-R to dissect the individual contributions of our framework's two core components: \textit{Semantic Exploration} (SE) and \textit{Activation Compaction} (AC). 
The results in Table~\ref{tab:ablation_study} not only confirm the necessity of both but also reveal their synergistic relationship.

Enabling only AC yields a modest improvement over the baseline, suggesting that while beneficial, simply compacting subspaces is insufficient if their semantic orientations are not actively managed, leading to potential collisions.
Conversely, the SE-only variant provides a more substantial performance leap, confirming that guiding subspace orientation is the primary mechanism for mitigating interference.
However, the performance gap to the full model indicates that without the compaction from AC, these correctly oriented subspaces remain unnecessarily diffuse, leaving less room for future tasks.

Ultimately, it is the synergistic interplay of both components that achieves the best result.
SE first steers activations into the correct "lanes," and AC then narrows these lanes to their essential width. 
This unequivocally demonstrates that managing both the \textbf{orientation} and \textbf{rank} of activation subspaces is indispensable for effectively mitigating catastrophic forgetting.

\subsection{Discussion}

\paragraph{Comparison of Activation Guidance and Parameter Constraints.}
We use t-SNE~\cite{maaten2008visualizing} to study the source of inter-task interference across adapter layers and tasks. Figure~\ref{fig:dist_down} shows that the down-projection parameters ($W_{\text{down}}$) from the same layer cluster closely across tasks, indicating that they mainly learn shared, task-agnostic representations. In contrast, the up-projection parameters ($W_{\text{up}}$) are widely spread across tasks in Figure~\ref{fig:dist_up}, indicating more task-specific representations and a greater risk of interference. Table~\ref{tab:discussion_param_vs_act} supports this finding: constraining only $W_{\text{up}}$ performs best among the parameter-based methods, whereas constraining the task-agnostic $W_{\text{down}}$ is less effective (see Appendix for implementation details). This agreement between the visualization and quantitative results identifies $W_{\text{up}}$ as a more relevant target for interference mitigation.

Figure~\ref{fig:dist_features} further shows that adapter input activations are general and similar across tasks in early ViT layers but become more task-specific in later layers. This layer-wise change explains why applying SGDS only to later layers performs best in Table~\ref{tab:discussion_param_vs_act}. Moreover, all SGDS variants outperform the strongest parameter-based method. These results show that directly guiding task-specific activation subspaces is more precise and effective for reducing interference than applying global constraints to the parameter space.

\begin{figure}[t]
    \centering
    \begin{subfigure}[b]{0.38\columnwidth}
        \centering
        \includegraphics[width=\textwidth]{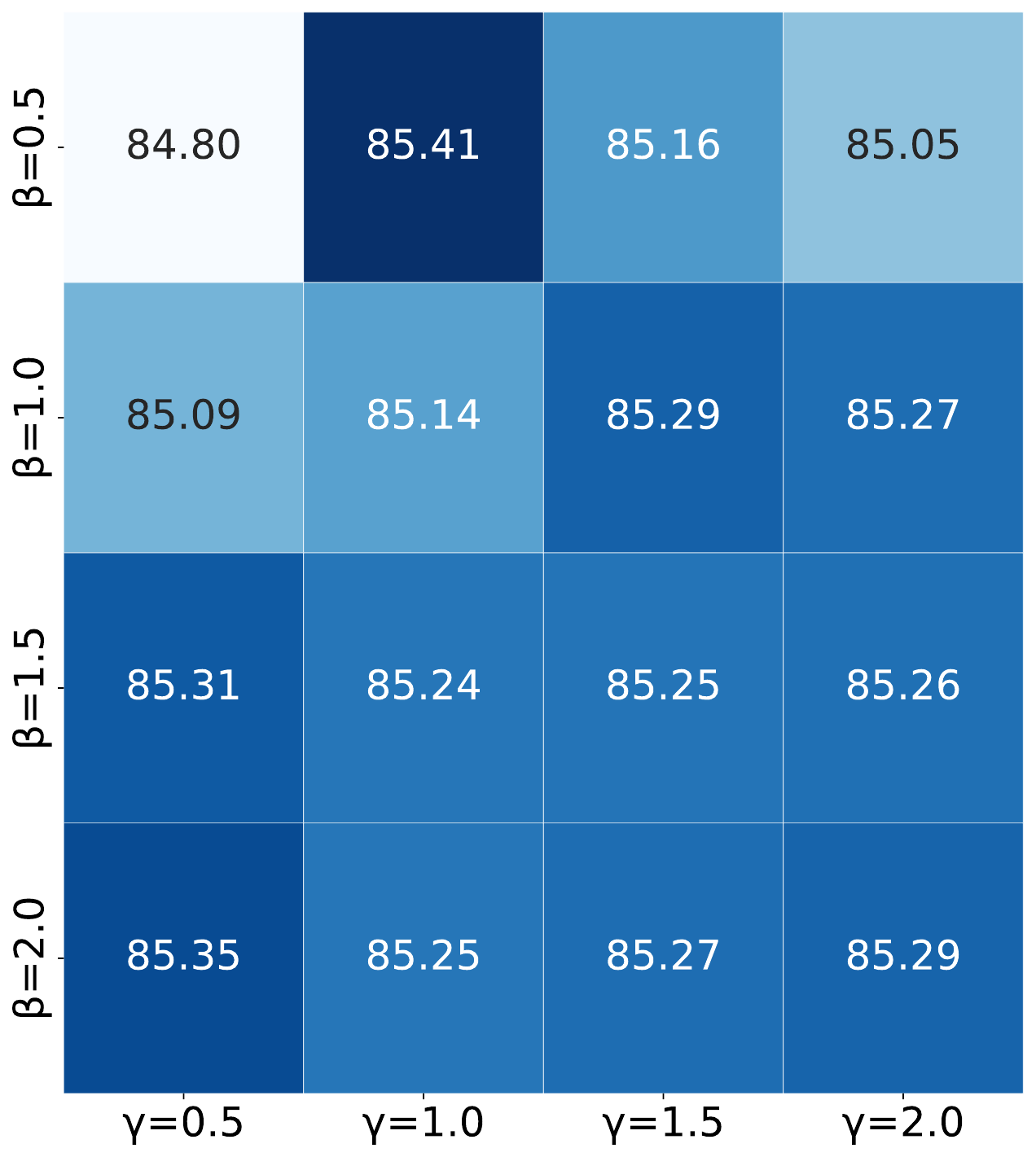} 
        \caption{Sensitivity to $\gamma$ and $\beta$.}
        \label{fig:sub_heatmap}
    \end{subfigure}
    \hfill 
    \begin{subfigure}[b]{0.58\columnwidth}
        \centering
        \includegraphics[width=\textwidth]{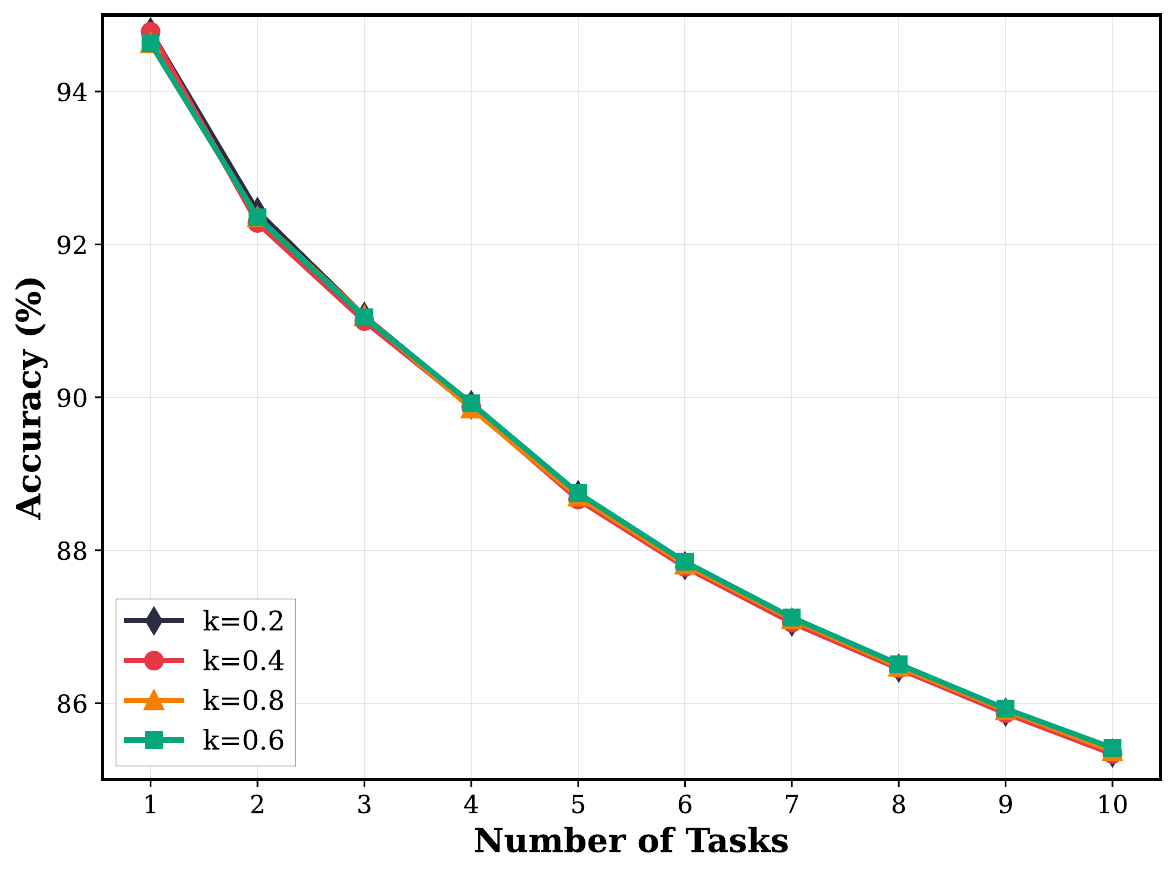} 
        \caption{Sensitivity to sparsity $k$.}
        \label{fig:sub_lineplot}
    \end{subfigure}
    
    \caption{
        Hyperparameter Robustness on ImageNet-R.
        The plots show the average incremental accuracy as a function of (a) the compaction strength $\gamma$ and exploration balance $\beta$, and (b) the sparsity ratio $k$. 
        SGDS demonstrates stable performance across a wide range of values for all key hyperparameters.
    }
    \label{fig:hyperparam_analysis}
\end{figure}

\paragraph{Parameter Robustness.}
SGDS is governed by three key hyperparameters: the compaction strength $\gamma$ (Eq.~\ref{eq:feat_consol}), the exploration balance $\beta$ (Eq.~\ref{eq:prob_nca}), and the sparsity ratio $k$. For all main experiments, we use a fixed setting of $\gamma=1.0$, $\beta=0.5$, and $k=0.6$. To demonstrate the method's robustness, we conduct a comprehensive sensitivity analysis on the ImageNet-R dataset, with results shown in Figure~\ref{fig:hyperparam_analysis}. The heatmap in Figure~\ref{fig:sub_heatmap} displays performance as we vary both $\gamma$ and $\beta$, showing consistently high accuracy. Similarly, Figure~\ref{fig:sub_lineplot} shows stable performance as the sparsity ratio $k$ is varied. This analysis confirms that SGDS is not critically dependent on precise hyperparameter tuning, underscoring its practical applicability.

\begin{figure}[t]
    \centering
    \begin{subfigure}[b]{0.48\columnwidth}
        \centering
        \includegraphics[width=\textwidth]{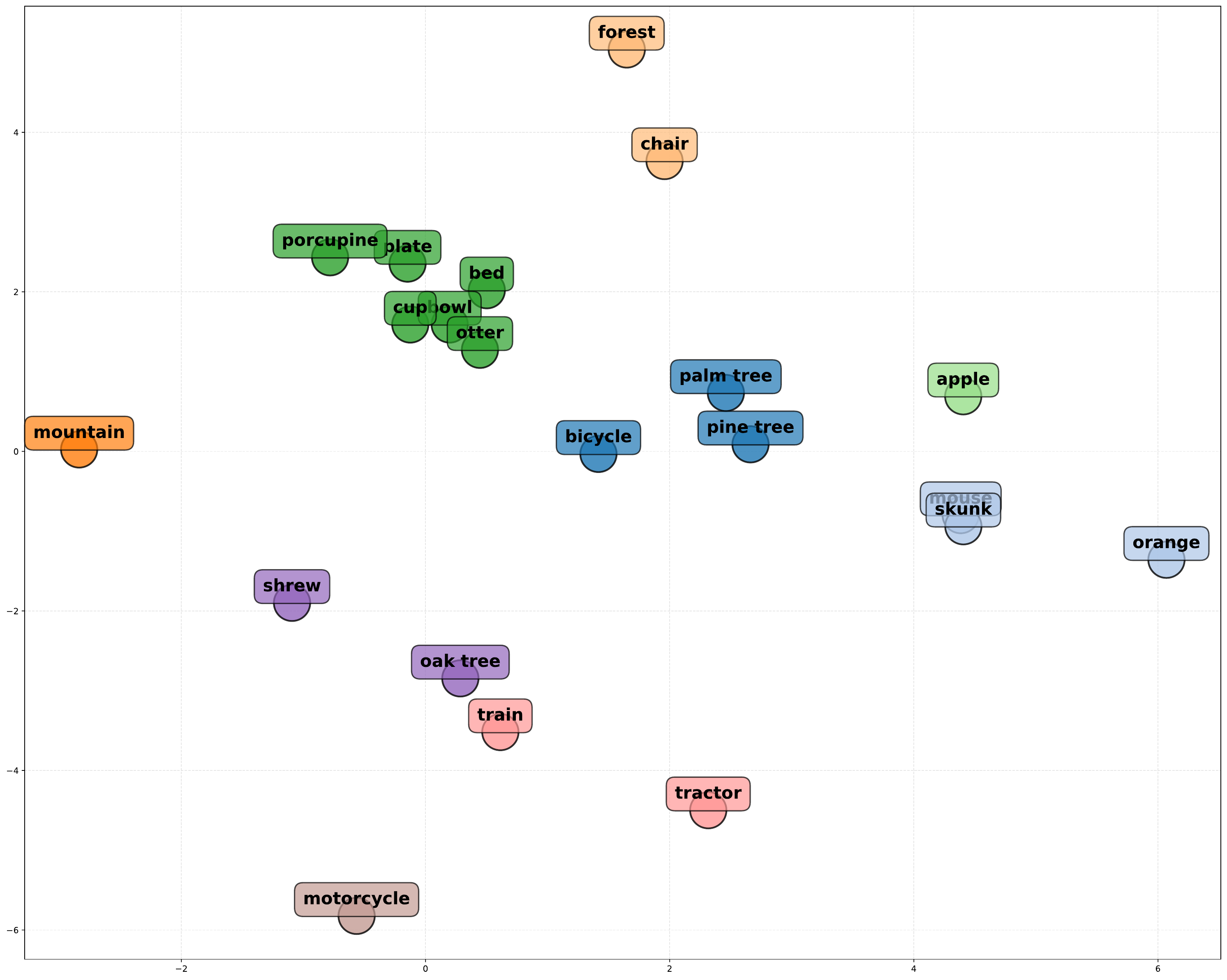} 
        \caption{Baseline (w/o SGDS)}
        \label{fig:tsne_baseline}
    \end{subfigure}
    \hfill 
    \begin{subfigure}[b]{0.48\columnwidth}
        \centering
        \includegraphics[width=\textwidth]{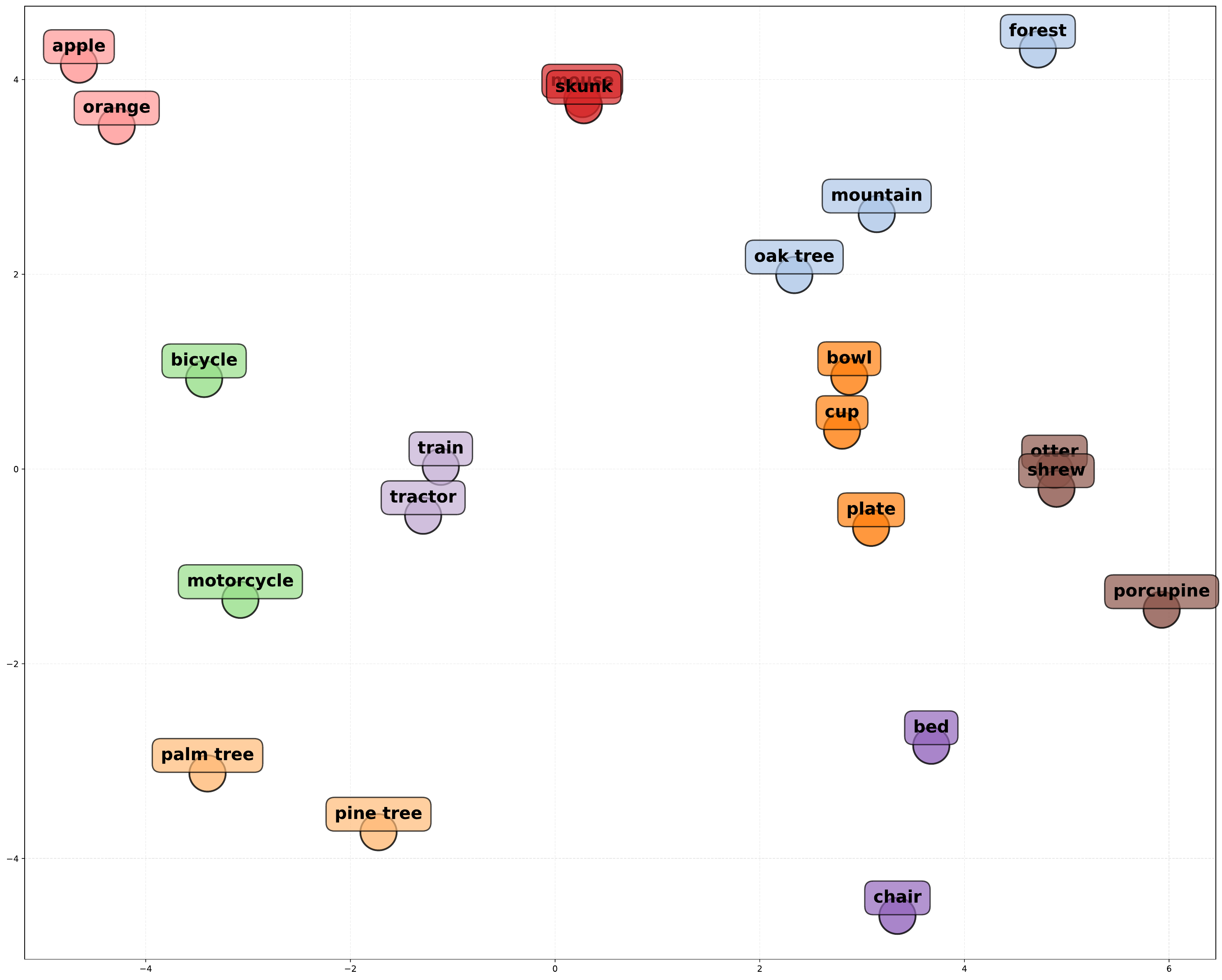} 
        \caption{Our Method (SGDS)}
        \label{fig:tsne_sgds}
    \end{subfigure}
    \caption{
        t-SNE Visualization of Activation Frequencies on CIFAR-100.
        The visualization shows activation frequencies from the final network layer.
        The baseline model (a) shows mixed activations with no clear structure.
        In contrast, SGDS (b) organizes activations into distinct and compact clusters, grouping semantically similar classes (e.g., vehicles, natural objects) together.
    }
    \label{fig:tsne_visualization}
\end{figure}

\paragraph{Visualizations.}
We use t-SNE~\cite{maaten2008visualizing} to visualize the distribution of input activation frequencies for the adapter in the final layer. Figure~\ref{fig:tsne_visualization} compares the activation space of a model trained with SGDS to one trained without it. As shown in Figure~\ref{fig:tsne_baseline}, the activations from the model without SGDS are largely intermingled, with no clear structure. In contrast, Figure~\ref{fig:tsne_sgds} shows that SGDS organizes the activation space into distinct clusters. Related classes are grouped together; for instance, fruit objects like 'apple' and 'orange' form one region, while vehicles such as 'tractor' and 'train' are co-located in another. Furthermore, the activations for each class occupy a tight, compact area. This visually demonstrates that SGDS constrains activations to low-rank subspaces. This well-structured activation space is the mechanism by which SGDS effectively mitigates inter-task interference.
\section{Conclusion}
\label{sec:conclusion}

Achieving a robust stability-plasticity balance remains a central challenge in Class-Incremental Learning.
While PTM-based adapters offer an efficient solution, their reliance on rigid parameter-space constraints to mitigate forgetting often compromises model plasticity.
To address this, we propose Semantic-Guided Dynamic Sparsification, a novel method that shifts the focus from constraining parameters to proactively guiding activations.
SGDS operates by governing both the orientation and rank of activation subspaces through targeted, dynamic sparsification.
This allows the model to construct sparse, task-specific subspaces that mitigate interference without imposing rigid constraints on the model's parameters.
Extensive experiments on major CIL benchmarks demonstrate that SGDS achieves superior accuracy over current SOTA methods, validating the effectiveness of activation-space guidance for achieving a more robust stability-plasticity balance. We discuss the limitations and future work in Appendix.

\bibliography{references}

\appendix
\onecolumn
\setcounter{table}{0}
\renewcommand{\thetable}{A\arabic{table}}
\setcounter{figure}{0}
\renewcommand{\thefigure}{A\arabic{figure}}

\section{Limitations and Future Work}
\label{app:limitations}
Despite the strong performance and efficiency of SGDS, we acknowledge certain limitations that offer avenues for future research.

First, similar to most standard CIL methods, our framework's semantic strategy decisions are made at discrete task boundaries. While this aligns with common CIL protocols, extending SGDS to a fully online or task-free setting, where data arrives as a continuous stream without explicit boundaries, remains a challenging but promising direction. This would require adapting our statistical counters and strategy formulation to operate in a more dynamic, sample-by-sample fashion.

Second, the effectiveness of our activation-guiding method is naturally linked to the richness of the feature space provided by the frozen pre-trained backbone. While we have demonstrated strong results on Vision Transformers, a crucial future step is to investigate how SGDS generalizes to different architectural paradigms, particularly the even larger and more complex backbones of emerging Foundation Models. 

Third, our semantic guidance currently relies on class-mean prototypes to determine the similarity between classes. This method, while effective and efficient, may not optimally capture the nuances of classes with high intra-class variance or multi-modal distributions. Future work could enhance the versatility of our framework by exploring more sophisticated semantic representations or by integrating knowledge from large-scale vision-language models for a more fine-grained semantic understanding.

We plan to address these aspects in future work to further enhance the robustness and autonomy of our CIL framework.

\section{Implementation Details of the Baseline method}
\label{app:baseline_details}

This appendix provides detailed implementations for the key mechanisms of the PTM-based CIL method introduced in Sec.~\ref{sec:methodology}'s Preliminaries. These details are based on state-of-the-art methods like TUNA~\cite{wang2025integrating} and serve as the foundation for our experimental comparisons.

\subsection{Inference-Time Adapter Retrieval}
\label{app:retrieval}
At inference time, since the task identity is unknown, the baseline method employs a sophisticated multi-stage mechanism to generate a prediction for an input sample $\mathbf{x}$.

\paragraph{Universal Adapter Fusion.}
First, a \textit{universal adapter} $\mathcal{A}_{\text{uni}}$ is created by merging all $t$ learned task-specific adapters $\{\mathcal{A}_1, \ldots, \mathcal{A}_t\}$ to capture shared cross-task knowledge. The weights of each adapter $\mathcal{A}_i$ are flattened into a vector $\mathbf{v}^i$. The universal adapter vector $\mathbf{v}^{\text{uni}}$, is then constructed by combining a consensus sign with a maximum magnitude. For each parameter at index $d$, this is formalized as:
\begin{equation}
    \mathbf{v}^{\text{uni}}[d] = \text{sign}\left(\sum_{i=1}^t \mathbf{v}^i[d]\right) \cdot \max_{i \in \{1, \ldots, t\}} \left|\mathbf{v}^i[d]\right|.
    \label{eq:universal_adapter_merge}
\end{equation}
The resulting vector $\mathbf{v}^{\text{uni}}$ is then reshaped to form the final universal adapter $\mathcal{A}_{\text{uni}}$.

\paragraph{Entropy-based Adapter Selection.}
Next, the most suitable task-specific adapter $\mathcal{A}_{t^*}$ is selected based on prediction confidence, measured by entropy. The model computes output probabilities for $\mathbf{x}$ using every available adapter $\mathcal{A}_i$. The adapter yielding the lowest entropy (i.e., most confident prediction) is chosen.
\begin{align}
\label{eq:app_min_entropy}
    \mathcal{A}_{t^*} = \underset{\mathcal{A}_i \in \{\mathcal{A}_1, \dots, \mathcal{A}_t\}}{\operatorname{argmin}} \Biggl( &-\sum_{c \in \mathcal{Y}_t} P(y=c|\mathbf{x}; \mathcal{A}_i) \nonumber \\
    &\times \log P(y=c|\mathbf{x}; \mathcal{A}_i) \Biggr).
\end{align}
Here, $P(y=c|\mathbf{x}; \mathcal{A}_i)$ is the softmax probability for class $c$ obtained using adapter $\mathcal{A}_i$.

\paragraph{Final Prediction Ensemble.}
Finally, the prediction is made by ensembling the output logits from the selected task-specific adapter $\mathcal{A}_{t^*}$ and the universal adapter $\mathcal{A}_{\text{uni}}$ to leverage both specific and shared knowledge:
\begin{equation}
\label{eq:app_prediction}
     y^* = \underset{c \in \mathcal{Y}_t}{\operatorname{argmax}} \left( \mathbf{w}_c^\top\phi(\mathbf{x};\mathcal{A}_{t^*}) + \mathbf{w}_c^\top\phi(\mathbf{x};\mathcal{A}_{\text{uni}}) \right).
\end{equation}

\subsection{Prototype-based Classifier and Alignment}
\label{app:alignment}
The classifier head $W$ is constructed non-parametrically from class prototypes. For each new class $c \in \mathcal{C}_t$, its prototype $\mathbf{p}_c$ is the mean feature vector of its training samples, extracted using the current task's adapter $\mathcal{A}_t$:
\begin{equation}
\label{eq:app_prototype_calc}
    \mathbf{p}_c = \frac{1}{|\mathcal{D}_c|} \sum_{x \in \mathcal{D}_c} \phi(x; \mathcal{A}_t).
\end{equation}
The classifier weight $\mathbf{w}_c$ for the new class is then its L2-normalized prototype: $\mathbf{w}_c = \mathbf{p}_c / \|\mathbf{p}_c\|_2$.

A critical challenge is the \textbf{subspace misalignment problem}, where prototypes from different tasks reside in incompatible feature subspaces. To address this misalignment without requiring old data, we adopt the strategy from MOS~\cite{sun2025mos}. This involves modeling the feature distribution of each past class $c'$ as a multivariate Gaussian, $\mathcal{N}(\boldsymbol{\mu}_{c'}, \boldsymbol{\Sigma}_{c'})$, using statistics stored from previous tasks.

To align the classifier, a set of pseudo-features is synthesized by sampling from these stored Gaussian distributions. Critically, these pseudo-features are \textbf{not} passed through the network again. Instead, they are used \textbf{directly} in the feature space to estimate the shift between the old and new subspaces. This estimated shift is then used to update the prototypes of old classes, effectively re-mapping them into the feature subspace of the current task. By combining these newly aligned prototypes for old classes with the new class prototypes, we construct a fully aligned classifier $W_t = [\mathbf{w}_1, \ldots, \mathbf{w}_{|\mathcal{Y}_t|}]$ for the current task.

\section{Algorithm Details}
\label{appendix:algorithm}

Algorithm~\ref{alg:sgds_manual} provides a comprehensive overview of the training process for our SGDS method.

\begin{algorithm*}[h!]
    \caption{SGDS Training}
    \label{alg:sgds_manual}
    \textbf{KwIn: }{Current task data $\mathcal{D}_t$, classes $\mathcal{C}_t$, past classes $\mathcal{Y}_{t-1}$, frozen backbone $\phi_{\text{frozen}}$, past adapters $\{\mathcal{A}_1, \ldots, \mathcal{A}_{t-1}\}$, global counter $\mathbf{F}$, class-specific counter $\mathbf{F}_c$, sparsity $k$, balance $\beta$, strength $\gamma$, target layers $\mathcal{L}_{\text{target}}$, epochs $E$.} \\
    \textbf{KwOut: }{Trained adapter $\mathcal{A}_t$.} \\
    
    \textcolor{gray}{\# --- Phase 1: Semantic Strategy Formulation ---} \\
    1. Compute prototypes $\{\boldsymbol{\mu}_y\}_{y \in \mathcal{Y}_{t-1} \cup \mathcal{C}_t}$. \\
    2. \textbf{For }{each new class $c \in \mathcal{C}_t$} \\
    {
        \hspace*{1em} a. Compute scores $S_{\text{old}}(c)$ and $S_{\text{new}}(c)$ via Eq.~\ref{eq:score_old} and Eq.~\ref{eq:score_new}. \\
        \hspace*{1em} b. \textbf{if }{$S_{\text{old}}(c) > S_{\text{new}}(c)$}, set $\text{strategy}[c] \leftarrow \text{KNOWLEDGE\_REUSE}$. \\
        \hspace*{1em} c. \textbf{else}, set $\text{strategy}[c] \leftarrow \text{NEW\_SUBSPACE\_ALLOCATION}$. \\
    }
    
    \textcolor{gray}{\# --- Phase 2: Two-Stage Adapter Training ---} \\
    3. Initialize new adapter $\mathcal{A}_t$ and expand classifier $W$. \\
    4. \textbf{For } {epoch from 1 to $E$} \\
    {
        \hspace*{2em} \textbf{For }{each batch $(x, y)$ in $\mathcal{D}_t$} \\
            \hspace*{4em} a. Initialize activations with input features: $\mathbf{a} \leftarrow x$. \\
            \hspace*{4em} b. \textbf{For }{each layer $l$ in target layers $\mathcal{L}_{\text{target}}$} \\
                \hspace*{6em} \textcolor{gray}{\# Two-stage sparsification based on epoch} \\
                \hspace*{6em} \textbf{if }{epoch $\le E/2$} \textcolor{gray}{\# Stage 1: Exploration} \\
                    \hspace*{8em} \textbf{if }{$\text{strategy}[y] == \text{KNOWLEDGE\_REUSE}$}, compute $\mathbf{p}^l$ via Eq.~\ref{eq:prob_kr}. \\
                    \hspace*{8em} \textbf{else}, compute $\mathbf{p}^l$ via Eq.~\ref{eq:prob_nca}. \\
                \hspace*{6em} \textbf{else} \textcolor{gray}{\# Stage 2: Compaction} \\
                    \hspace*{8em} Compute probabilities $\mathbf{p}^l$ via Eq.~\ref{eq:feat_consol}. \\
                \hspace*{6em} Generate probabilistic mask: $\mathbf{m}_{prob}^l \sim \text{Bernoulli}(\mathbf{p}^l)$. \\
                \hspace*{6em} Apply probabilistic mask: $\mathbf{a}'_l \leftarrow \mathbf{a}_l \odot \mathbf{m}_{prob}^l$. \\
                \hspace*{6em} Apply deterministic selection: $\mathbf{a}_l \leftarrow \text{TopK}(\mathbf{a}'_l, \lfloor k \cdot N \rfloor)$. \\
                \hspace*{6em} \textcolor{gray}{\# Update historical counters for the final selected activations} \\
                \hspace*{6em} Find indices $\mathcal{I}_{final} \leftarrow \text{Indices}(\mathbf{a}_l \neq 0)$. \\
                \hspace*{6em} \textbf{For }{each index $j \in \mathcal{I}_{final}$} \\
                    \hspace*{8em} $\mathbf{F}[l, j] \mathrel{+}= 1$ and $\mathbf{F}_c[y, l, j] \mathrel{+}= 1$. \\
            \hspace*{2em} c. Pass final activations through network to get logits. \\
            \hspace*{2em} d. Compute classification loss $\mathcal{L}_{\text{cls}}$. \\
            \hspace*{2em} e. Update parameters of $\mathcal{A}_t$ via backpropagation. \\
    }
    
    \textcolor{gray}{\# --- Phase 3: Classifier Update ---} \\
    5. Update classifier $W$ for new classes using class-mean prototypes. \\
    
    \textbf{Return} {Trained adapter $\mathcal{A}_t$.}
\end{algorithm*}
\begin{figure*}[t!]
    \centering
    \begin{subfigure}[b]{0.49\textwidth}
        \centering
        \includegraphics[width=\textwidth]{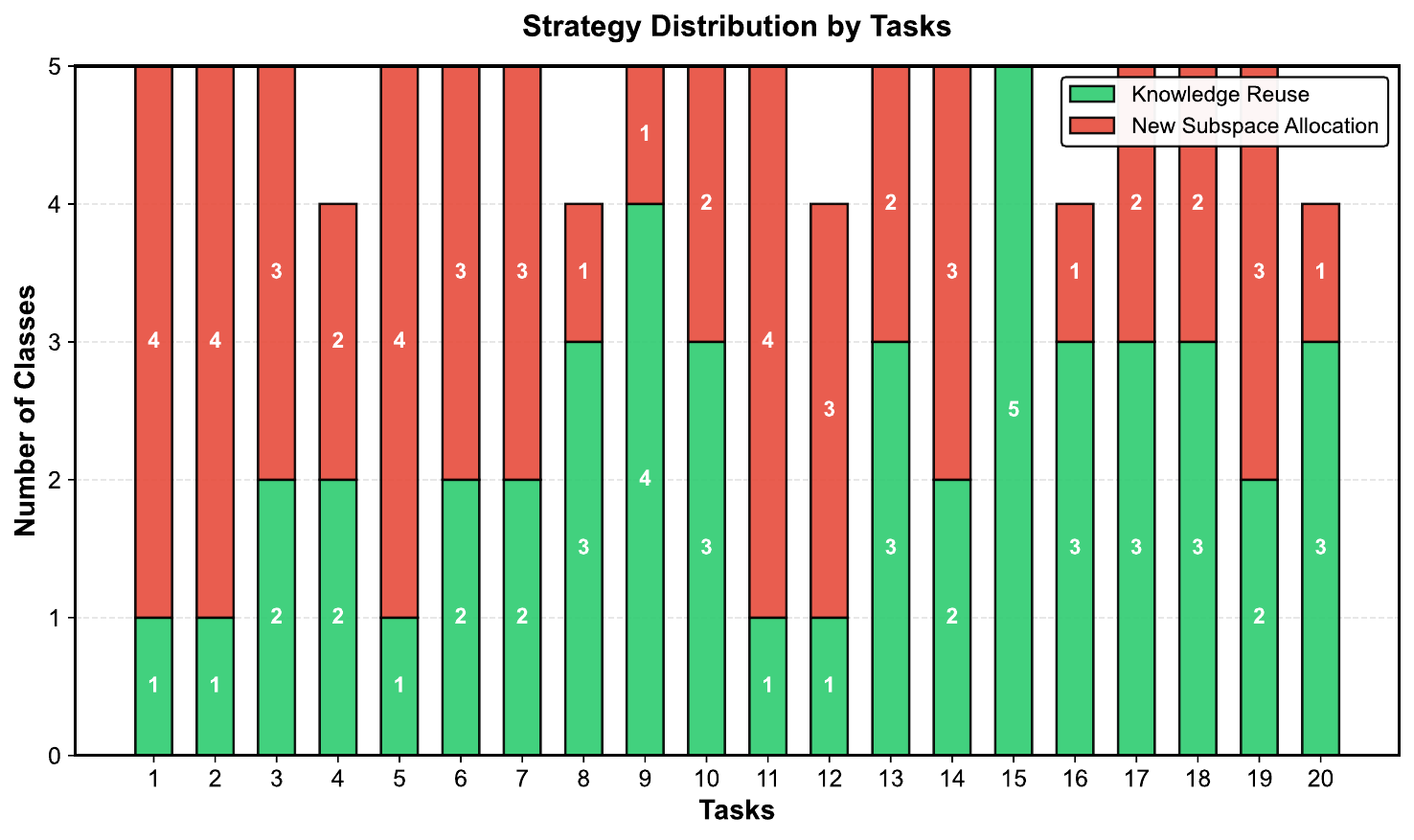}
        \caption{Strategy distribution per task.}
        \label{fig:strategy_dist}
    \end{subfigure}
    \hfill 
    \begin{subfigure}[b]{0.49\textwidth}
        \centering
        \includegraphics[width=\textwidth]{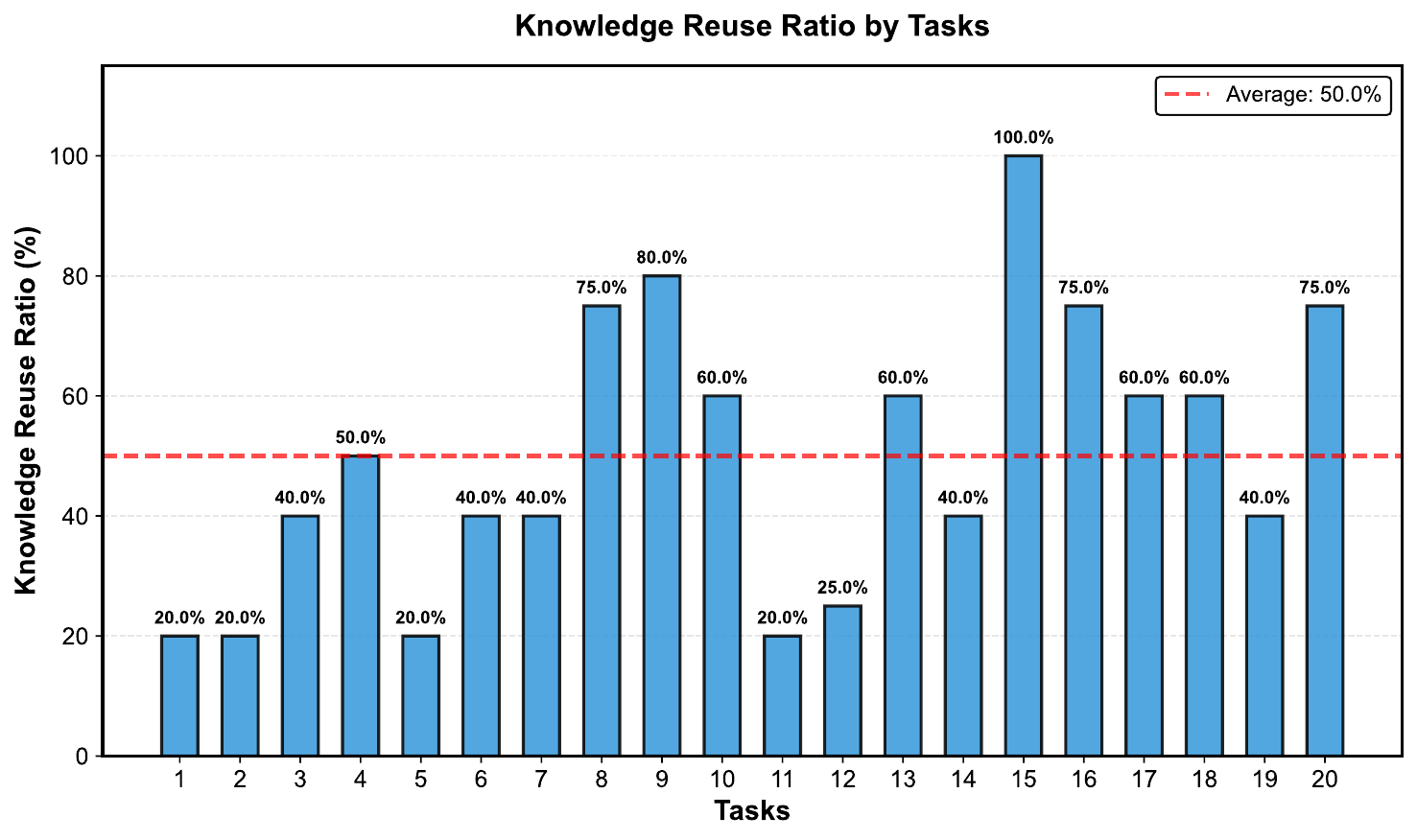}
        \caption{Knowledge reuse ratio.}
        \label{fig:reuse_ratio}
    \end{subfigure}
    \caption{
        Dynamic Adaptation of Learning Strategies in a Long Task Sequence (CIFAR-100, 20 Tasks).
        This figure illustrates how SGDS dynamically adjusts its learning strategy.
        \textbf{(a)} The left panel shows the per-task count of classes assigned to \textit{Knowledge Reuse} versus \textit{New Subspace Allocation}. Early tasks (e.g., 1-5) are dominated by \textit{New Subspace Allocation}, demonstrating high plasticity as the model builds its initial semantic foundation. Later tasks show an increasing prevalence of \textit{Knowledge Reuse}.
        \textbf{(b)} The right panel tracks the ratio of classes designated for \textit{Knowledge Reuse}. The ratio steadily increases, indicating that as the model's knowledge base expands, it increasingly and efficiently leverages existing subspaces. This automatic shift from a plasticity-dominant to a stability-focused strategy is crucial for SGDS's strong performance in long-sequence CIL scenarios.
    }
    \label{fig:long_sequence_analysis}
\end{figure*}
\begin{figure}[t!]
    \centering
    \includegraphics[width=0.5\linewidth]{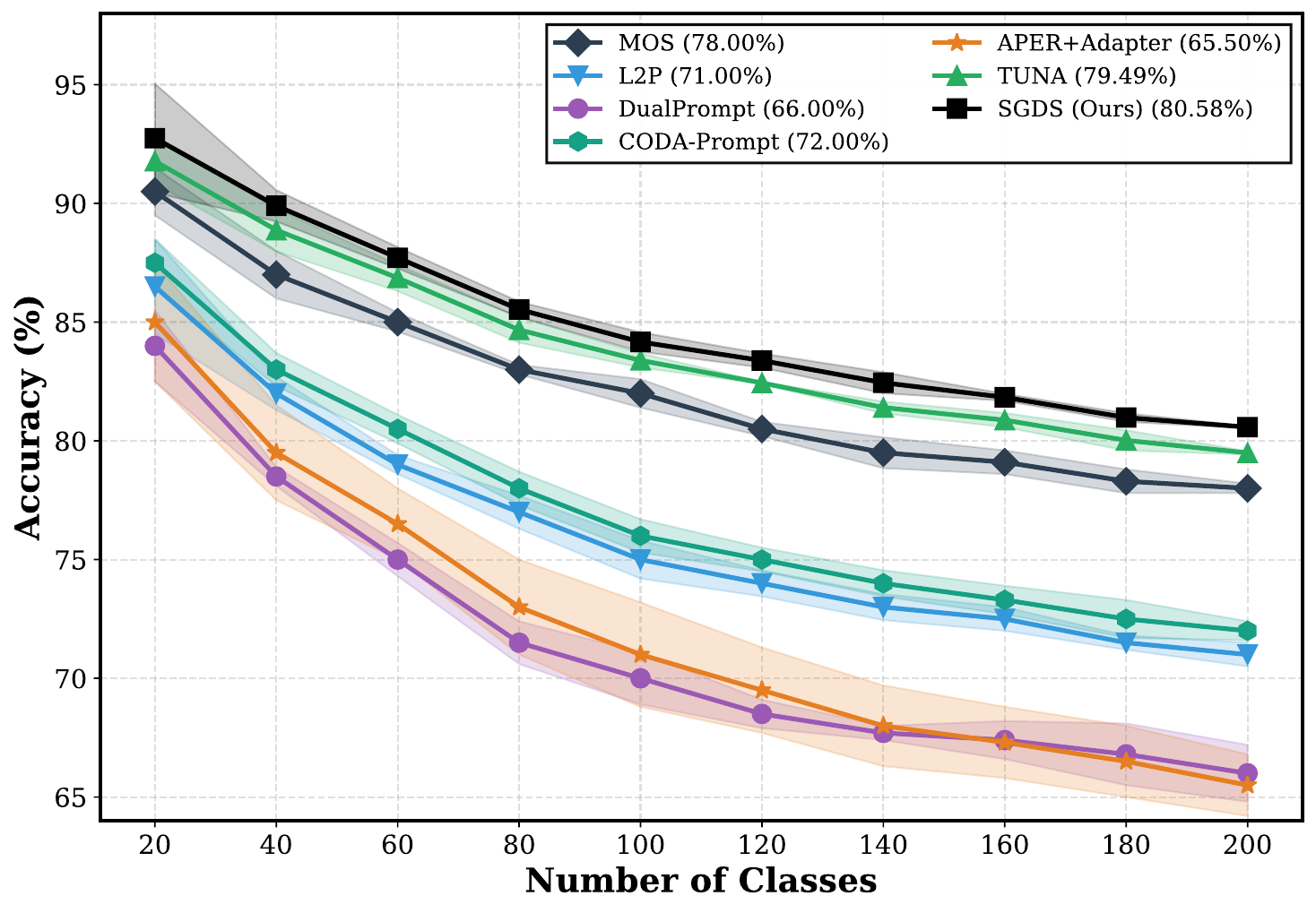} 
    \caption{
        Performance stability across multiple random seeds on ImageNet-R (10-task setting).
        The plot displays the average accuracy as the number of learned tasks increases. 
        All results shown are the mean performance calculated over five different random seeds, which vary the order of incoming classes. 
        The consistent and superior performance of SGDS highlights its robustness against variations in the incremental learning sequence.
    }
    \label{fig:multiple_runs}
\end{figure}

\section{Full Implementation Details}
\label{app:implementation_details}

\noindent\textbf{Datasets and Task Protocols.}
The benchmarks include standard data and domain shifts, with ImageNet-A and ObjectNet used to test robustness to domain shifts. CIFAR-100 has 100 classes and is split into 20 tasks with 5 classes each. ImageNet-R, ImageNet-A, and ObjectNet each have 200 classes and are split into 10 tasks with 20 classes each. Following~\cite{rebuffi2017icarl}, we shuffle the class orders with a fixed random seed of 1993 before task partitioning.

\noindent\textbf{Training Details.}
We implement all experiments in PyTorch~\cite{paszke2019pytorch} based on the PILOT codebase~\cite{sun2023pilot} and run them on NVIDIA GeForce RTX 2080 Ti GPUs. For a fair comparison, all methods use the same ViT-B/16 backbone pre-trained on ImageNet-21K. For each incremental task, we train the models for 20 epochs with a batch size of 48. We use SGD with momentum, an initial learning rate of 0.01, and a cosine annealing schedule. SGDS uses 60\% global activation sparsity only in the final layer of the ViT backbone. We set the exploration balance $\beta$ (Eq.~\ref{eq:prob_nca}) to 0.5 and the compaction strength $\gamma$ (Eq.~\ref{eq:feat_consol}) to 1.0. For all adapter-based methods, including ours, the intermediate projection dimension is 16.

\section{Implementation of Parameter-Space Regularization}
\label{app:param_reg}

To provide a comprehensive comparison against parameter-centric methods, we implemented three variants of orthogonality regularization for the adapter modules, as analyzed in our main discussion (Section~\ref{sec:experiments}). The total loss function for these experiments is a combination of the standard cross-entropy loss and a regularization term:
\begin{equation}
\mathcal{L} = \mathcal{L}_{\text{cls}} + \lambda \mathcal{L}_{\text{reg}},
\end{equation}
where $\lambda$ is a hyperparameter balancing the two terms. The regularization loss, $\mathcal{L}_{\text{reg}}$, is defined based on the Frobenius norm of the dot product between the adapter matrices of the current task $t$ and all previous tasks $i \in \{1, \dots, t-1\}$. We evaluated the following three strategies:

\begin{itemize}
    \item \textbf{Regularize $W_{\text{up}}$ only:} The regularization is applied exclusively to the up-projection matrices.
    \begin{equation}
        \mathcal{L}_{\text{reg}} = \sum_{l \in \mathcal{L}_{\text{mask}}} \sum_{i=1}^{t-1}\left\| {W_{\text{up}}^{(l,t)}} \cdot  {(W_{\text{up}}^{(l,i)})}^\top \right\|_{F}^2
    \end{equation}
    
    \item \textbf{Regularize $W_{\text{down}}$ only:} The regularization is applied exclusively to the down-projection matrices.
    \begin{equation}
        \mathcal{L}_{\text{reg}} = \sum_{l \in \mathcal{L}_{\text{mask}}} \sum_{i=1}^{t-1}\left\| {W_{\text{down}}^{(l,t)}} \cdot  {(W_{\text{down}}^{(l,i)})}^\top \right\|_{F}^2
    \end{equation}
    
    \item \textbf{Regularize Both:} The regularization loss is the sum of the losses from both up- and down-projection matrices.
\end{itemize}

\section{Performance Stability Across Multiple Runs}
\label{appendix:multiple_runs}

In the main paper, our experiments followed the standard protocol from Rebuffi et al.~\cite{rebuffi2017icarl}, using a fixed random seed of 1993 to determine the class order for each incremental task. To rigorously assess the robustness of our method against variations in task composition, this section extends our analysis by repeating the experiments on ImageNet-R with five different random seeds: \{1993, 1994, 1995, 1996, 1997\}.

This procedure generates five distinct result sets for each evaluated method. The mean performance for each method is visualized in Figure~\ref{fig:multiple_runs}. As the figure illustrates, SGDS consistently outperforms all competing methods across the different random seeds, confirming the stability and general effectiveness of our activation-space guidance method, irrespective of the specific class ordering.

\begin{figure*}[t!]
    \centering
    \begin{subfigure}[b]{0.48\textwidth}
        \centering
        \includegraphics[width=\textwidth]{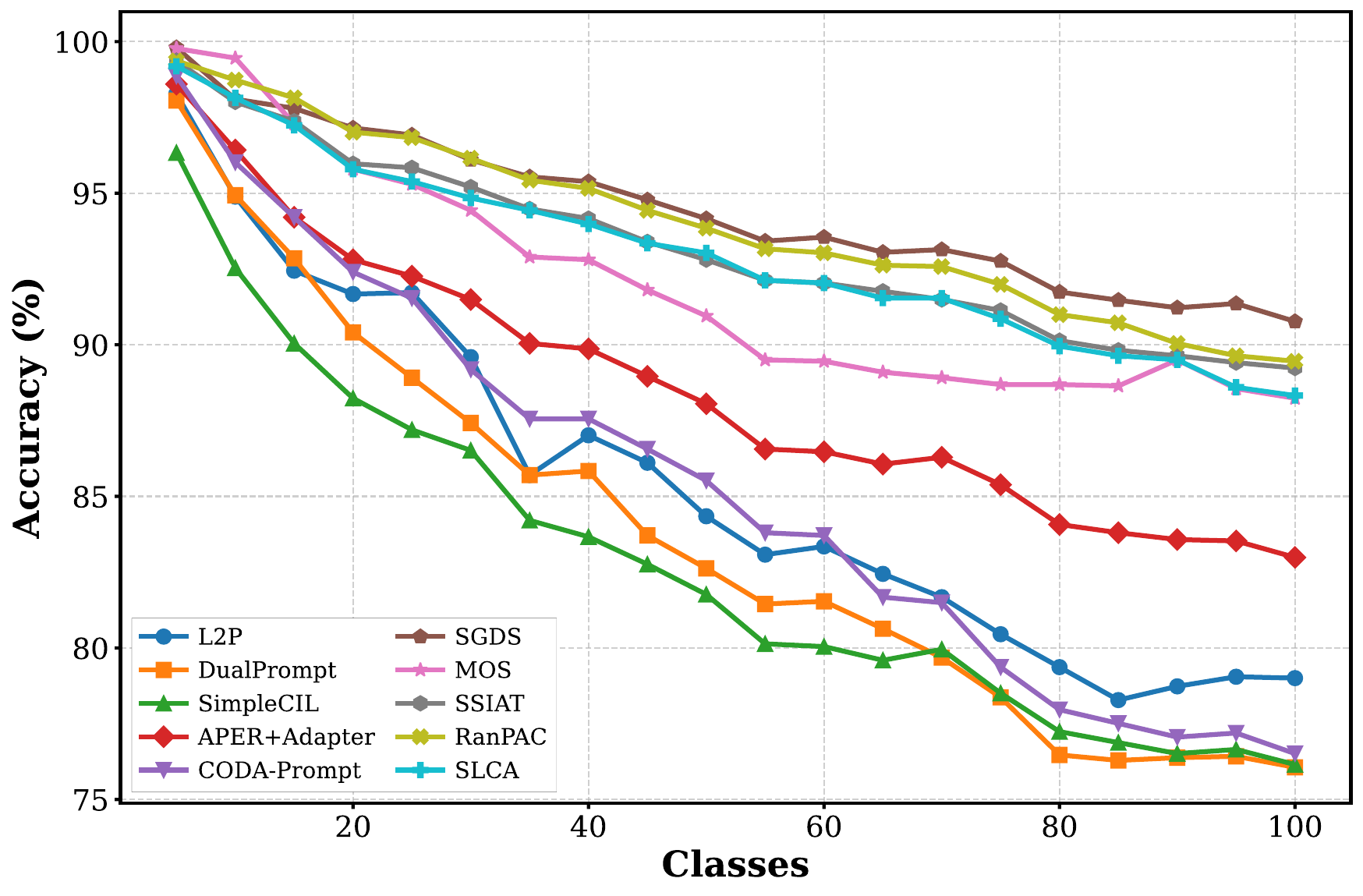}
        \caption{CIFAR-100}
        \label{subfig:curve_cifar100}
    \end{subfigure}
    \hfill
    \begin{subfigure}[b]{0.48\textwidth}
        \centering
        \includegraphics[width=\textwidth]{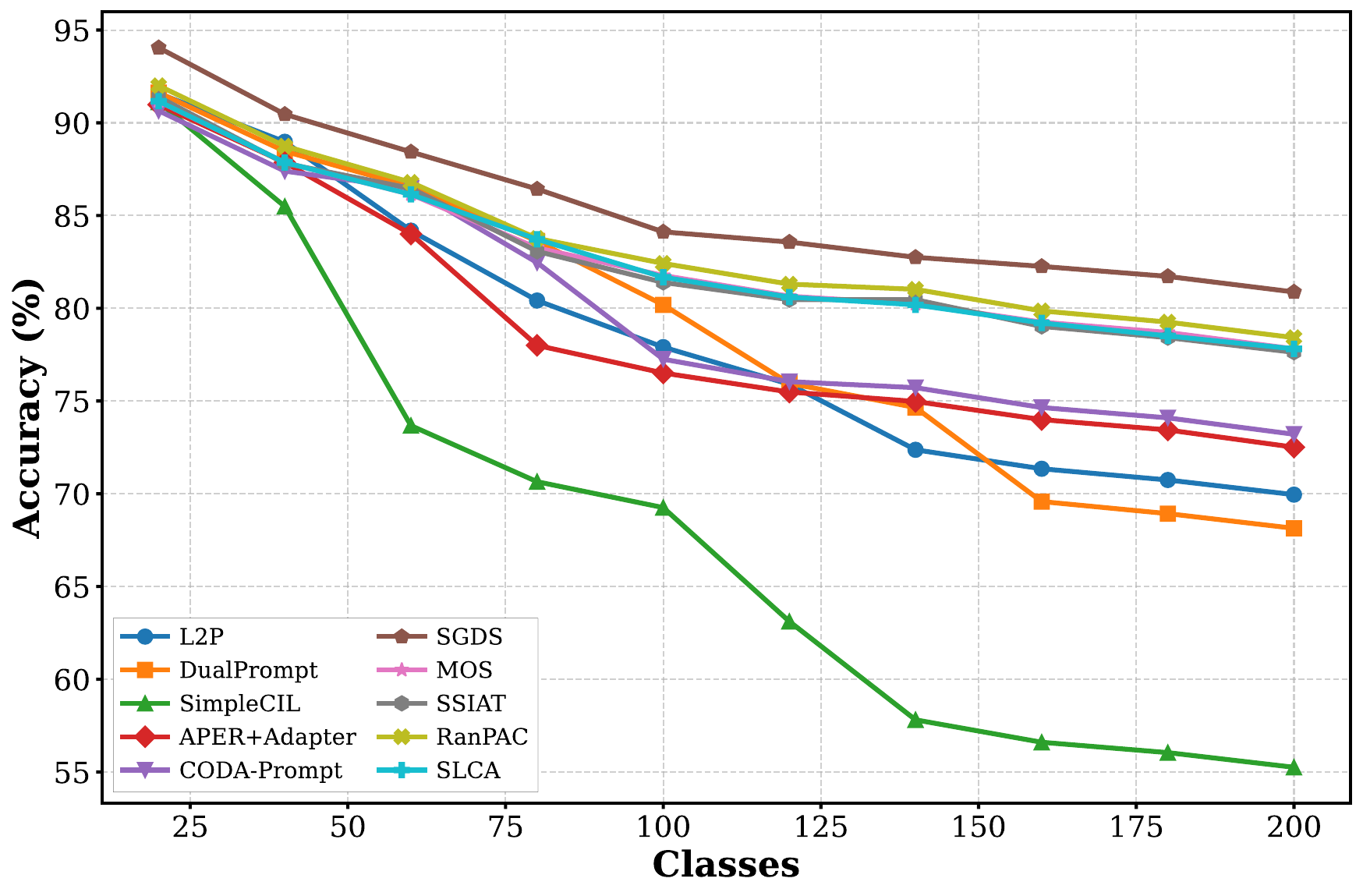}
        \caption{ImageNet-R}
        \label{subfig:curve_imagenetr}
    \end{subfigure}

    \vspace{0.5cm} 

    \begin{subfigure}[b]{0.48\textwidth}
        \centering
        \includegraphics[width=\textwidth]{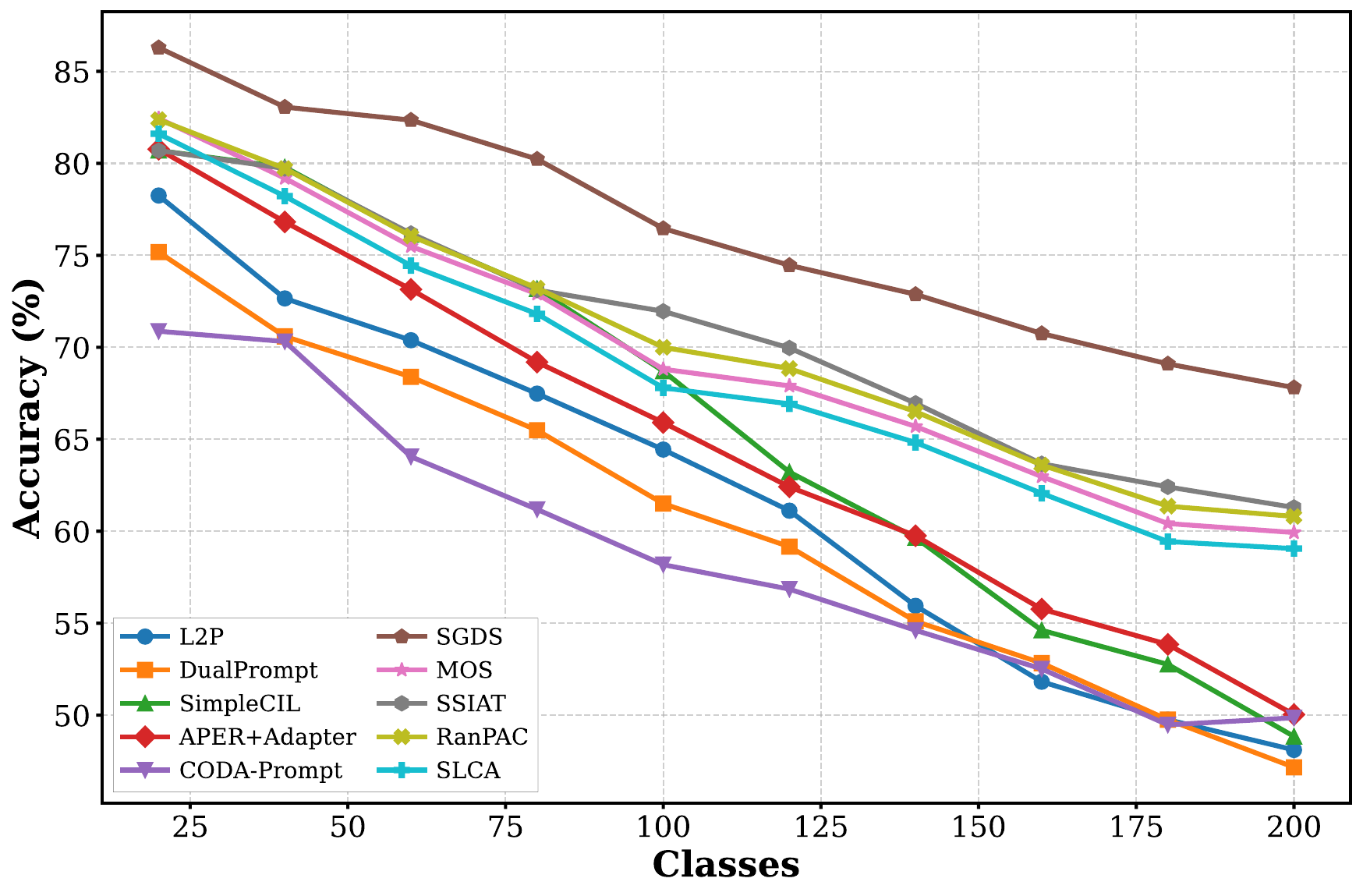}
        \caption{ImageNet-A}
        \label{subfig:curve_imageneta}
    \end{subfigure}
    \hfill
    \begin{subfigure}[b]{0.48\textwidth}
        \centering
        \includegraphics[width=\textwidth]{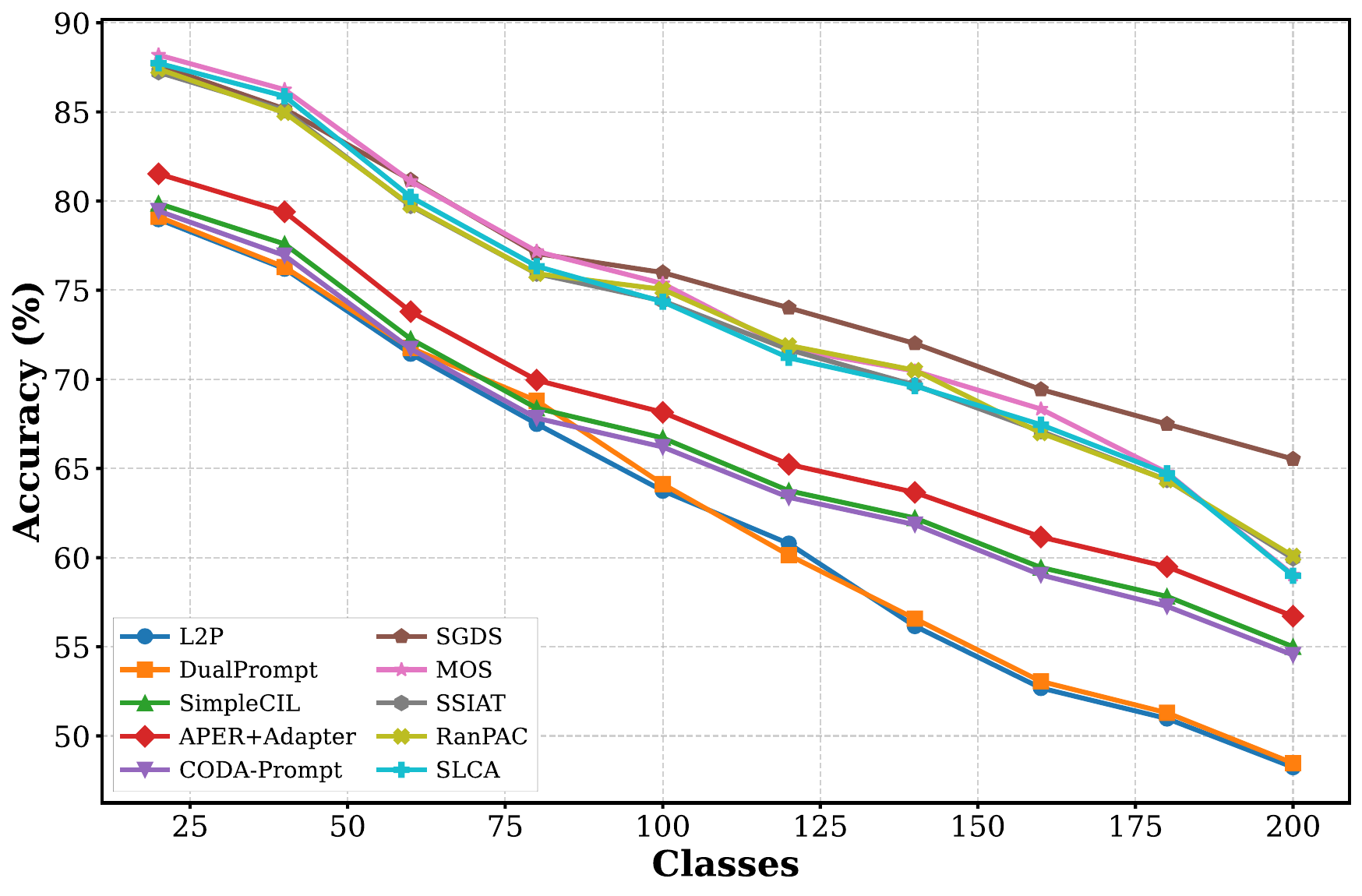}
        \caption{ObjectNet}
        \label{subfig:curve_objectnet}
    \end{subfigure}

    \caption{
        Incremental accuracy trends of SGDS compared to SOTA CIL methods on four standard benchmarks: CIFAR-100, ImageNet-R, ImageNet-A, and ObjectNet. Across all scenarios, SGDS demonstrates consistently superior performance. This highlights its robust ability to mitigate catastrophic forgetting and maintain a clear advantage as the number of learned classes increases.
    }
    \label{fig:appendix_curves}
\end{figure*}
\section{Analysis of Learning Strategy in Long Task Sequences}
\label{appendix:long_sequence}

To demonstrate the adaptive nature of SGDS in long-sequence scenarios, we analyze the distribution of its two learning strategies: \textit{Knowledge Reuse} versus \textit{New Subspace Allocation}, across a 20-task CIFAR-100 experiment. This analysis reveals how our method automatically balances plasticity and stability over an extended learning horizon, which is critical for effective lifelong learning.

As shown in Figure~\ref{fig:long_sequence_analysis}, the strategy distribution dynamically adapts throughout the learning process. In the initial tasks, where the model's learned knowledge is sparse, SGDS predominantly allocates new, orthogonal subspaces to accommodate novel semantics (\textit{New Subspace Allocation}). This prioritizes plasticity, allowing the model to learn effectively without the constraint of a pre-existing structure.

As the sequence progresses, the model accumulates a rich library of semantic subspaces. Consequently, SGDS increasingly identifies opportunities to map new classes to existing subspaces (\textit{Knowledge Reuse}), as reflected by the rising reuse ratio shown in Figure~\ref{fig:reuse_ratio}. This strategic shift towards stability prevents representational drift and solidifies prior knowledge, enabling robust performance over extended learning horizons. This self-regulating mechanism is a key factor behind the superior performance of SGDS in long-sequence benchmarks.
\section{Detailed Incremental Performance Analysis}
\label{appendix:performance_curves}

To further analyze the performance progression of SGDS, we present the incremental accuracy trends against leading CIL methods in Figure~\ref{fig:appendix_curves}. 
These plots, based on a ViT-B/16 backbone pre-trained on ImageNet-1K, track the final accuracy as the number of learned classes increases across our four primary benchmarks.
Across all evaluated settings, the plots clearly show that SGDS not only achieves the highest final accuracy but also maintains a consistent and often widening performance margin throughout the entire learning sequence.
For instance, on the challenging ObjectNet sequence (Figure~\ref{subfig:curve_objectnet}), the performance of competing methods degrades more sharply over time, whereas SGDS maintains a more stable and superior trajectory.  
Unlike methods that may suffer from accumulating interference, SGDS's ability to construct well-structured activation subspaces ensures superior performance from the first task to the last.

\end{document}